\documentclass[sigconf, nonacm=true, balance=true]{acmart}

\AtBeginDocument{%
  \providecommand\BibTeX{{%
    \normalfont B\kern-0.5em{\scshape i\kern-0.25em b}\kern-0.8em\TeX}}}

\makeatletter
\@namedef{ver@lineno.sty}{9999/12/31}
\@namedef{opt@lineno.sty}{}
\makeatother

\usepackage{graphicx}
\usepackage{balance}
\usepackage{xcolor}
\usepackage{xargs}                      
\usepackage{xparse}
\usepackage[colorinlistoftodos,prependcaption]{todonotes}
\definecolor{atomictangerine}{rgb}{1.0, 0.6, 0.4}
\definecolor{columbiablue}{rgb}{0.61, 0.87, 1.0}
\definecolor{lava}{rgb}{0.81, 0.06, 0.13}
\usepackage[frozencache,cachedir=listings]{minted}
\usepackage[most, minted]{tcolorbox}

\usepackage{subcaption}

\usepackage{stfloats}

\newcommandx{\unsure}[2][1=]{\todo[linecolor=atomictangerine,backgroundcolor=atomictangerine!25,bordercolor=atomictangerine,#1]{#2}}
\newcommandx{\fix}[2][1=]{\todo[linecolor=lava,backgroundcolor=lava!25,bordercolor=lava,#1]{#2}}
\newcommandx{\info}[2][1=]{\todo[linecolor=columbiablue,backgroundcolor=columbiablue!25,bordercolor=columbiablue,#1]{#2}}
\newcommandx{\improvement}[2][1=]{\todo[linecolor=Plum,backgroundcolor=Plum!25,bordercolor=Plum,#1]{#2}}
\newtcblisting{openhps}{%
	listing engine=minted,
	colback=gray!5,			
	minted language=minted/lexers/typescript.py:TypeScriptLexer -x,
	minted style=vs,
	listing only,
	enhanced,
	boxrule=1pt,			
	colframe=gray!20,		
	minted options = {
		linenos, 			
		breaklines=false,	
		fontsize=\footnotesize, 
		numbersep=2mm,		
		tabsize=2,			
		stripall=true		
	},
	overlay={
		\begin{tcbclipinterior}
			\fill[gray!20] (frame.south west) rectangle ([xshift=4mm]frame.north west);
		\end{tcbclipinterior}
	},
	before upper={
		\footnotesize
		\mintinline{typescript}{/* @openhps/core | version 0.2.0 */}
	}
}

\setcopyright{none} 

\begin{document}

\title{OpenHPS: An Open Source Hybrid Positioning System}


\author{Maxim Van de Wynckel}
\orcid{https://orcid.org/0000-0003-0314-7107}
\affiliation{%
	\institution{Web \& Information Systems Engineering Lab \\
		Vrije Universiteit Brussel}
	\streetaddress{Pleinlaan 2}
	\city{1050 Brussels}
	\country{Belgium}
}
\email{mvdewync@vub.be}

\author{Beat Signer}
\orcid{https://orcid.org/0000-0001-9916-0837}
\affiliation{%
	\institution{Web \& Information Systems Engineering Lab \\
		Vrije Universiteit Brussel}
	\streetaddress{Pleinlaan 2}
	\city{1050 Brussels}
	\country{Belgium}
}
\email{bsigner@vub.be}


\renewcommand{\shortauthors}{Maxim Van de Wynckel and Beat Signer}

\begin{abstract}
  Positioning systems and frameworks use various techniques to determine the position of an object. Some of the existing solutions combine different sensory data at the time of positioning in order to compute more accurate positions by reducing the error introduced by the used individual positioning techniques. We present OpenHPS, a generic hybrid positioning system implemented in TypeScript, that can not only reduce the error during tracking by fusing different sensory data based on different algorithms, but also also make use of combined tracking techniques when calibrating or training the system. In addition to a detailed discussion of the architecture, features and implementation of the extensible open source OpenHPS framework, we illustrate the use of our solution in a demonstrator application fusing different positioning techniques. While OpenHPS offers a number of positioning techniques, future extensions might integrate new positioning methods or algorithms and support additional levels of abstraction including symbolic locations.
\end{abstract}

\keywords{OpenHPS; hybrid positioning; open source; processing network; indoor positioning}



\maketitle

\section{Introduction}
Determining the location of a person or asset is an important aspect of various human-computer interaction~(HCI) solutions. Position tracking can be used to create autonomous vehicles, navigation systems and in context brokers such as CoBrA~\cite{chen2003ontology} to create implicit interactions based on the location and possibly other contextual information~\cite{schmidt1999there}. While we mainly rely on the Global Positioning System~(GPS) to determine our location outdoors, many other positioning techniques exist that work both, indoors as well outdoors and sometimes even have been designed for completely different use cases as described later in Section~\ref{subsec:positioning-methods}. Each positioning technique has its advantages and disadvantages depending on the environment where it is being used. Hybrid positioning systems use the combination of different techniques and sensors to determine a more accurate position through sensor fusion~\cite{chen2015fusion}.

One of the disadvantages with most commonly used indoor positioning methods such as Bluetooth Beacons or Wi-Fi access point positioning is the requirement of some form of calibration or training. With our proposed hybrid system called \emph{OpenHPS}\footnote{\href{https://www.openhps.org}{https://www.openhps.org}}, the goal is to achieve hybrid positioning during both, the navigation or tracking (\emph{online stage}) and between tracking, training and calibration of the used positioning techniques (\emph{offline stage}). Unlike many of the existing frameworks, the goal of OpenHPS is to offer a layered abstraction supporting a wide range of positioning techniques and fusion algorithms.

Using our knowledge on various positioning techniques and algorithms which are discussed in Section~\ref{subsec:positioning-methods}~and Section~\ref{subsec:positioning-algorithms}, we identified and analysed the actors of our system as highlighted in Section~\ref{subsec:actors}. This analysis revealed that actors play a different role depending on the positioning method being used. In order to allow the system to combine different existing positioning methods, we opted for a processing stream network where each node in a graph topology contributes to the sampled data. 

Our design goal with the chosen processing network is to handle positioning data in real time, giving developers complete control over the data flow. The proposed OpenHPS framework is presented in Section~\ref{sec:solution} together with an illustrative demonstrator use case before discussing some future work in Section~\ref{sec:conclusion}. 

\section{Background}
\label{sec:background}

The first step in conceptualising our hybrid positioning framework was to investigate existing positioning methods in order to find their similarities as well as differences. OpenHPS has to be able to support a wide range of different positioning methods and implementation goals. When analysing existing approaches, we made a distinction between positioning \emph{methods} and \emph{algorithms}, similar to Wilson~et~al.~\cite{sakpere2017state}. Positioning methods represent the techniques and technologies that are available to determine a location while positioning algorithms include the algorithms that can be used to combine these methods.

\subsection{Positioning Methods} \label{subsec:positioning-methods}
In this section we list some of the more prominent positioning methods ranging from signal-based to visual solutions. Note that the selected positioning methods represent a limited set of the techniques and functional requirements we aim to support.

\subsubsection{Global Position System (GPS)} \label{subsubsec:gps}
Starting with the most well-known technology, the Global Positioning System~(GPS) is used for outdoor positioning~\cite{bulusu2000gps}. It makes use of satellites in an orbit around the Earth to triangulate a location consisting of a longitude, latitude and elevation. There also exist variations on the original global position system, including Differential~GPS~(DGPS)~\cite{rezaei2007kalman} or the Assisted~GPS~(a-GPS)~\cite{djuknic2001geolocation} that fuses GPS with dead reckoning described later in Section~\ref{subsubsec:dead-reckoning}.

\subsubsection{RF-based Positioning} \label{subsubsec:rf-based-positioning}
RF-based positioning is a commonly used technique for indoor environments. Examples of technologies used for RF-based positioning include Wi-Fi, Bluetooth Beacons, RFID and LTE~cell towers. These RF~signals offer a \emph{landmark} that can be used as a reference when determining the position based on fingerprinting or other mathematical calculations.

\subsubsection{Simultaneous Localisation and Mapping (SLAM)} \label{subsubsec:slam}
Simultaneous Localisation and Mapping, or SLAM for short, is a positioning method that makes use of 2D~sensors to map its surroundings. One of the more common examples includes the use of LIDAR (Light Detection and Ranging)~\cite{mendes2016icp}, capturing distance readings around a sensor. These readings can then be used to generate a 2D~map of the environment. 

\subsubsection{Visual Positioning Techniques}\label{subsubsec:visual}
~~Existing~~~visual~~~positioning techniques use image sensors to determine the position of the object the sensor is attached to (e.g.~Visual SLAM~\cite{openvslam2019}) or the position of objects in its field of view. In previously mentioned positioning methods, the \emph{tracked} object obtains the sensor data. With Multi-Target Multi-Camera Tracking~(MTMCT)~\cite{krumm2000multi} the tracked object is moving within the field of view of one or more image sensors.

\subsubsection{Dead Reckoning} \label{subsubsec:dead-reckoning}
Dead reckoning calculates the current position based on the previous known position and the velocity or acceleration that is applied to that position~\cite{holzke2019improving,beauregard2006pedestrian}. While the accuracy of dead reckoning is not ideal, it can be used to improved other positioning techniques such as GPS.

\subsection{Positioning Algorithms} \label{subsec:positioning-algorithms}
In order to calculate a position or to combine multiple positioning methods, different algorithms are used. The list of algorithms presented in this section offers a baseline for the types of positioning algorithms to be supported, but our OpenHPS framework is not limited to the presented set of algorithms.

\subsubsection{Triangulation and Multilateration} \label{subsubsec:triangulation-trilateration}
Mathematical operations such as triangulation and multilateration can be used for several positioning methods and technologies discussed in Section~\ref{subsec:positioning-methods}. For techniques that provide a landmark such as RF-based positioning, the received signal strength~(RSS) might provide a rough estimate of the distance. Other examples include mathematical positioning using a time difference with respect to the time of arrival~(ToA) or the angle of arrival~(AoA).

\subsubsection{Fingerprinting} \label{subsubsec:rss-fingerprinting}
While with multilateration we only need information about the position of the used landmarks, fingerprinting requires a calibration for all possible positions in the tracking area~\cite{xiao2011integrated}. A fingerprint of the sensor data at a given provided position is created during the offline stage. Later, these stored fingerprints are used during the online stage to reverse the sensor data into a position.

\subsubsection{Noise Filtering} \label{subsubsec:noise-filtering}
Sensor data should be filtered, which can be done through different noise filters. Similar to dead reckoning, a noise filter often requires knowledge of previous sensor readings and positions to predict the next result.

Noise filtering is one of the main requirements of our hybrid positioning system. The reason why we want to combine multiple technologies or algorithms is to reduce errors and noise filtering is the key component in realising this error reduction of positioning data. Note that individual positioning methods such as object recognition or dead reckoning may want to perform different types of noise filtering algorithms tailored to their data.

\subsubsection{Machine Learning} \label{subsubsec:machine-learning}
This type of algorithms includes a number of machine learning algorithms that can be of aid during calibration as well as positioning. These algorithms require training during the offline stage with the results being deployed during the online stage. This issue will be further discussed when discussing the requirements in Section~\ref{subsec:requirements}, where we allow data to be used in the online and offline stages of the OpenHPS framework.

\subsubsection{Computer Vision}
With the visual positioning methods discussed in Section~\ref{subsubsec:visual}, the \emph{tracked} actor is not always uniquely identified. Such visual positioning methods have to be able to detect and track objects between multiple frames, camera angles or positions. The algorithms used to track and detect persons or objects from a video stream are beyond the scope of our framework, but OpenHPS should be able to provide a generic interface supporting these types of algorithms.

\subsection{Hybrid Positioning} \label{subsec:hybrid-positioning}
Apart from supporting different positioning methods and algorithms, OpenHPS should be able to combine these methods. This requires a choice of algorithms to specify how the result of each method can be used in the combined output.

Sensor fusion can occur at a low or high level~\cite{elmenreich2002introduction}. Raw sensor data such as IMU~sensors or the relative signal strength from a transmitter can be fused in noise filtering algorithms. On a high level, calculated or provided positions (i.e.~by third-party positioning systems) with a certain predicted accuracy can be combined using linear regression, heuristic weighted averages or any \emph{decision fusion} algorithm. 

\section{Related Work} \label{sec:related-work}
Location-based Services~(LBS) represent a generalised category of systems that provide the current location of a person or other objects~\cite{kupper2005location}. A distinction between a push- and pull-based LBS is made~\cite{steiniger2006foundations}. A pull LBS provides a location when it is requested to do so, while a push LBS delivers information when a new location is determined by a provider.

The idea of combining multiple positioning methods in an LBS is not new. In this section we present some related work, ranging from existing hybrid positioning systems, frameworks, their used terminology and standards throughout location-based services. 

\subsection{Location APIs and Specifications} \label{subsec:specifications}
On the Web, the Geolocation~API~\cite{popescu2010geolocation} offers a high-level interface for single or repeated position updates. The API provides an abstraction of the underlying technologies and algorithms used to determine the position. However, developers can request a high accuracy result or maximum cache age if hardware permits this. Resulting positions are geographical coordinates complying with the WGS84 standard~\cite{decker1986world}.

JSR-179 and the improved JSR-293~\cite{barbeau2008location} specifications are Java 2 Micro Edition~(J2ME) modules that provide developers an API to obtain the location and orientation of a mobile device. Included in the API is a storage interface for landmarks~(see Section~\ref{subsubsec:rf-based-positioning}). The specification represents locations as timestamped coordinates with an orientation, accuracy, speed and information about the used positioning method~\cite{di2005indoor}. When requesting a location, criteria such as the desired accuracy, power consumption and response timeout can be provided. 

WebXR~\cite{jones2019webxr,maclntyre2018thoughts} is a Web~API that provides an interface for the tracking and use of VR~or~AR~headsets. The API uses the \emph{pose} terminology to indicate the position and orientation of the person wearing an XR~headset in 3D~space. While WebXR should not be considered as a location API, its specification uses terminology that is common in our framework. As an API that provides a tracking position, it adds a goal for our framework to support these third-party APIs.

\subsection{Hybrid Positioning Systems}
Various research concerning the fusion of sensor data to predict a more accurate position exists. SignalSLAM~\cite{mirowski2013signalslam} represents an example of a hybrid system that uses signals of various positioning methods such as GPS, Wi-Fi and Bluetooth to map the surroundings. Chen~et~al.~\cite{chen2015fusion} have shown how a smartphone can combine sensor data of Wi-Fi access point positioning and Pedestrian Dead Reckoning~(PDR). This combination of dead reckoning with another positioning method is a common combination used by many hybrid systems. LearnLoc~\cite{pasricha2015learnloc} is a smartphone-centred positioning framework that uses fingerprinting algorithms (k-nearest neighbours algorithm) in combination with various sensor data available on a smartphone to provide power-efficient indoor positioning. The consideration of power efficiency is a common requirement in mobile positioning systems. Other than achieving the most accurate position, these location-based services use sensor fusion to prevent the continuous use of precise sensors such as cameras or GPS.

IndoorAtlas provides a Platform-as-a-Service~(PaaS) with a well-established Software Development Kit~(SDK) for combining \mbox{Wi-Fi}, GPS, Bluetooth beacons, dead reckoning and even geomagnetic positioning~\cite{haverinen2014utilizing}. While the latter method has been found to be less ideal in steel reinforced buildings~\cite{pasricha2015learnloc}, it still offers a useful addition for creating a hybrid positioning system where geomagnetic positioning might be combined with other positioning methods.

In the research by Bekkelien~and~Deriaz~\cite{bekkelien2012hybrid} a framework called Global Positioning Module~(GPM) had been presented for in- and outdoor positioning. GPM provides a uniform interface to different position \emph{providers}. These providers are fused in a \emph{kernel} that selects the position based on provided \emph{criteria} (e.g.~precision, accuracy or detection probability). Their approach offers a clear methodology on how this criteria can contribute to the selection or fusion of different technologies. However, the position providers and kernels are implemented on a high level of abstraction providing no room for developers to choose different algorithms or fusion techniques.

Ficco~and~Russo~\cite{ficco2009hybrid} presented a technology-independent hybrid positioning middleware called HyLocSys. Position \emph{estimators}, representing different technologies, provide positions when a user performs a \emph{pull} of their current position. Sensor fusion combines these estimated positions into a final response. With the middleware being an extension of the JSR-179 specification presented earlier, these pull requests accept criteria such as the preferred response time and expected accuracy. Other than many frameworks that only provide geographical positions, HyLocSys provides geometric, symbolic as well as hybrid location models. Symbolic locations represent abstract places such as buildings, floors and rooms that are relatively positioned to each other. A hybrid location can convert this symbolic location to a geometric position. Note that the paper does not discuss positioning technologies such as dead reckoning or SLAM that require periodic updates in order to keep an up-to-date position.

Scholl~et~al.~\cite{scholl2012fast} propose a system that uses a LIDAR scanner to determine the fingerprinting position. This is somewhat similar to our goal of using different positioning methods to support the offline stage. 

The Robot Operating System~(ROS)~\cite{quigley2009ros} is a structured communication layer that can be used to create autonomous robots. It focuses on the integration of various robotics aspects such as positioning, computing and hardware interfacing. ROS provides the concept of peer-connected nodes that perform computational tasks. These nodes represent interchangeable software modules that help to build a pipeline from sensory data to an output action. For positions and orientations, ROS uses the \emph{pose} concept which contains both the position and orientation of a user.

Our framework should adhere to specifications such as WGS84 when working with geographical positions. However, unlike many of the related work discussed in this section, we also want to support non-geometric positions. The hybrid location model presented by Ficco~and~Russo~\cite{ficco2009hybrid} offers a good type of location, but is still heavily focused on geometric positions.

Positioning methods and algorithms are often represented under the term \emph{providers} that are optionally combined via high-level decision fusion. In our framework, we want to separate providers into generic algorithms and positioning methods that can easily be switched. This not only allows for more extensibility, but also some low-level sensor fusion.

The Geolocation~API, JSR-179 and HyLocSys allow for the specification of accuracy or other criteria when requesting a position. However, unlike high-level~APIs that hide the underlying technologies, OpenHPS is aimed towards developers with an understanding of the available hardware and positioning techniques that influence the criteria. 

The landmark storage of JSR-179 is a very useful addition to a positioning system, as it is a common requirement for many positioning techniques. The persistence of landmarks between the online and offline stage is an important requirement that is extended to fingerprinting information and cached position storage in OpenHPS. This persistence should allow us to interface with existing systems such as the Geolocation~API that support both push-based position updates as well as retrieving the current (cached) position.

\section{OpenHPS Framework} \label{sec:solution}
In this section, we present and discuss the system design of our proposed OpenHPS hybrid positioning framework. After listing some general requirements in Section~\ref{subsec:requirements}, we outline the overall architecture in Section~\ref{subsec:architecture}. Next, we provide some general information on our chosen implementation and demonstrate the use of OpenHPS in Section~\ref{subsec:implementation}~and~Section~\ref{subsec:demonstrator}.

\subsection{Requirements} \label{subsec:requirements}
Based on existing positioning methods and algorithms discussed in Section~\ref{subsec:positioning-methods} and Section~\ref{subsec:positioning-algorithms}, the following framework requirements have been derived. We start by specifying the actors of our system and motivate the use of a processing network where each node of the graph topology might represent one of these actors.

\subsubsection{System Actors}
\label{subsec:actors}
After investigating different existing positioning techniques, we defined four actors in OpenHPS: 

\begin{itemize}
	\item \textbf{Tracked actor}: This is an actor that can be tracked during the online positioning stage. A \emph{tracked actor} can be an end user or an asset that might optionally contain sensors to further support the tracking. Our main goal is to determine the most accurate position of this type of actor.
	\item \textbf{Tracking actor}: This type of actor is responsible for tracking a tracked actor. Note that for some positioning methods, the same actor might act as a \emph{tracking actor} as well as a tracked actor. However, for positioning methods such as the visual object tracking introduced in Section~\ref{subsubsec:visual}, the tracking actor is represented by the camera while a tracked actor is the object that is being detected.
	\item \textbf{Calibration actor}: Some positioning methods require a calibration before the positioning method can be used. Unlike the tracked actor, the purpose of a calibration actor is to train and calibrate how the tracking actor will be used in the online stage of the system.
	\item \textbf{Computing actor}: The computing actor is responsible for providing the final position output by our system. This actor combines the data generated by one or more tracking actors about a tracked actor and processes the data by, for example, using one of the positioning algorithms described in Section~\ref{subsec:positioning-algorithms}.
\end{itemize}

These four actors represent the four main components within OpenHPS. By distinguishing between the \emph{tracked} and \emph{tracking} actor, the system is able to support the tracking of persons or objects that do not actively participate in the positioning process.

\subsubsection{Functional Requirements} \label{subsubsec:functional-requirements}
In the following, we list the minimal functional requirements for our OpenHPS framework.

\begin{itemize}
	\item \textbf{Online stage positioning}: In order to perform hybrid positioning or sensor fusion, multiple (processed) sources need to be combined by using different algorithms.
	\item \textbf{Offline stage positioning}: Processed results can be used to calibrate positioning methods of another (online) stage.
	\item \textbf{Third-party frameworks}: Our framework needs to support third-party high-level positioning systems. These external systems might provide their own calculated position of a tracked actor that needs to be fused with the position determined by our framework. In addition, the identification of this tracked actor might differ between frameworks.
	\item \textbf{Environment mapping}: With the requirement to support positioning methods such as SLAM and VSLAM, the system does not only offer the possibility to output an absolute position, but might also create an environment map. Our solution should be capable of handling, storing and using this map to its advantage.
	\item \textbf{Decentralisation}: Our positioning framework should be able to combine the four different actors introduced in the previous section based on remote hardware. This requires the framework to work decentralised without requiring any centralised sensor fusion, which can be achieved by allowing multiple computing actors to work independently. However, developers should still be given the option to centralise certain parts of the system if needed.
	\item \textbf{Monotonicity:} Partial information from a source should result in a partial output. In the context of a positioning system, this means that a computing actor does not need the sensor data of all positioning methods to determine a position. This requirement also helps in the decentralisation and parallelisation of the framework.
\end{itemize}

\subsubsection{Non-functional Requirements} \label{subsubsec:nonfunctional-requirements}
The following non-functional requirements contributed to the final decision about the software language used for OpenHPS.
\begin{itemize}
	\item \textbf{Availability}: Our solution has to be available on various platforms ranging from servers to embedded systems; also supporting the decentralisation functional requirement.
	\item \textbf{Performance and latency}: Throughput is an important criteria when processing streaming data. Input data such as video and audio streams needs to be processed in real time. The latency also indicates how long it takes for data to be used in computations. As our goal is to achieve an accurate current position, outdated sensor data is not relevant.
	\item \textbf{Modularity}: The framework should be modular with both, a low-level API and modules that can be added and removed based on the available sensors and concrete use cases. Developers should remain in control of the types of algorithms and the flow of data from producer to consumer.
\end{itemize}

\subsection{Framework Architecture} \label{subsec:architecture}
In order to support the presented functional and non-functional requirements, we decided to build on a \emph{stream-based positioning system} that takes various types of \emph{input data} and processes this data to get the desired output. Data that is transmitted between nodes is encapsulated in so-called \emph{data frames} that can contain sensory data as well as one or more \emph{data objects} the sensor data applies to, and are described in detail in Section~\ref{subsubsec:data-frame}.

For the design of our process network, a number of existing stream- and layer-based frameworks such as Akka Streams~\cite{davis2019akka} or \mbox{TensorFlow}~\cite{abadi2016tensorflow} have been investigated. These frameworks solve similar issues and are further detailed in Section~\ref{subsubsec:graph-design}. Due to the fact that each node needs to be configured individually, the decision was made to investigate flow-based frameworks where each component of the stream network is added individually.

Unlike low-level data stream frameworks, OpenHPS focuses on data that is helpful for positioning. We offer a higher-level~API for creating the network and data that is handled by the sytem. Concepts such as edges or ports that are often found in stream-based programming languages are abstracted and not directly accessible by developers. However, unlike other hybrid frameworks~\cite{haverinen2014utilizing,bekkelien2012hybrid}, the stream processing is extensible enough to give developers the opportunity to modify the positioning methods along with the used algorithms.

We start by discussing our process network design that uses a graph topology similar to other stream frameworks. Next, we present the data frames, objects and positional data that are being handled by the network.

\subsubsection{Process Network Design}
\label{subsubsec:graph-design}
The OpenHPS~framework uses a process network to handle data. The data is processed and dynamically manipulated by multiple connected nodes in a predefined graph topology. In the following, we list our three main design goals for this network:

\begin{enumerate}
	\item \textbf{Consistent data types}: Data that is being processed in the network should have a reliable type and content. We process \mintinline{typescript}{DataObject}s encapsulated in \mintinline{typescript}{DataFrame}s, which provides a defined scope how generic parts of our network should handle information.
	\item \textbf{Processing goal}: Processing has the goal of providing an absolute position for our tracked actors. With this goal we have a clear understanding how every computing actor contributes to the output.
	\item \textbf{Producer priority}: The producer or tracking actor has the highest priority. Slow consumers or computing actors must not result in outdated sensor information. Rather, developers should be given the opportunity to control what happens with the overflow of information that cannot be processed timely.
\end{enumerate}

Starting from the goal of producing up-to-date positioning information, we opted for a push-pull-based stream. Data can be dropped if its not relevant for determining a more up-to-date position. The monotonicity of our framework ensures that positions can be determined based on partial data.

Each node can be designed to accept both push and pull requests. Similar to reactive streams~\cite{geilen2004reactive}, push and pull actions are \emph{promise}-based and can be executed asynchronously. If a node that receives a pull request cannot respond with a data frame itself, it will forward the pull request to its incoming node(s).

Different to a traditional pull that returns a \emph{response}, we use the \emph{push} terminology to indicate a response for a given pull. This behaviour and terminology is similar to Akka Streams~\cite{davis2019akka}, but unlike reactive streams where data can only be provided when there is a demand, there is no back pressure built into the stream itself. Using the push terminology for a pull response removes the ambiguity of a response arriving after an already existing push in the pipeline. It also enforces the design goal of producers having the highest priority, even if a producer only generates information when requested.

A regular node has a unique identifier and push/pull functionality for data frames. Each node can have $0\ldots{}n$ inlets or outlets. Our system consist of the following three subtypes of the regular node:

\begin{itemize}
	\item \textbf{Source node}: A source node provides a specific data type. This can either be a push or pull node that pushes data frames when they are available (e.g.~a camera recording at a fixed frame rate) or creates a new data frame when the downstream node asks for it via a pull request (e.g.~triggering a Bluetooth scan). The source node merges \emph{data objects} in the data frame with those that were previously stored via \emph{data services}. This merging behaviour prevents the need for feedback loops to gain knowledge on previously calculated positioning data.
	\item \textbf{Processing node}: A processing node is a higher-level interface for a regular node. It provides an abstraction on the push and pull functionality to simplify the creation of a processing function of either data frames or individual data objects.
	\item \textbf{Sink node}: An output node or sink node accepts a specific data type as output frame. Unlike processing nodes, this type of node will not push data to other nodes. Upon receiving a data frame, the data objects will be stored using a compatible data service. Once saved, an event is sent upstream to indicate that the processing of this frame and its contained objects is completed.
\end{itemize}

\begin{figure}[htb] 
	\centering
	\includegraphics[width=\columnwidth]{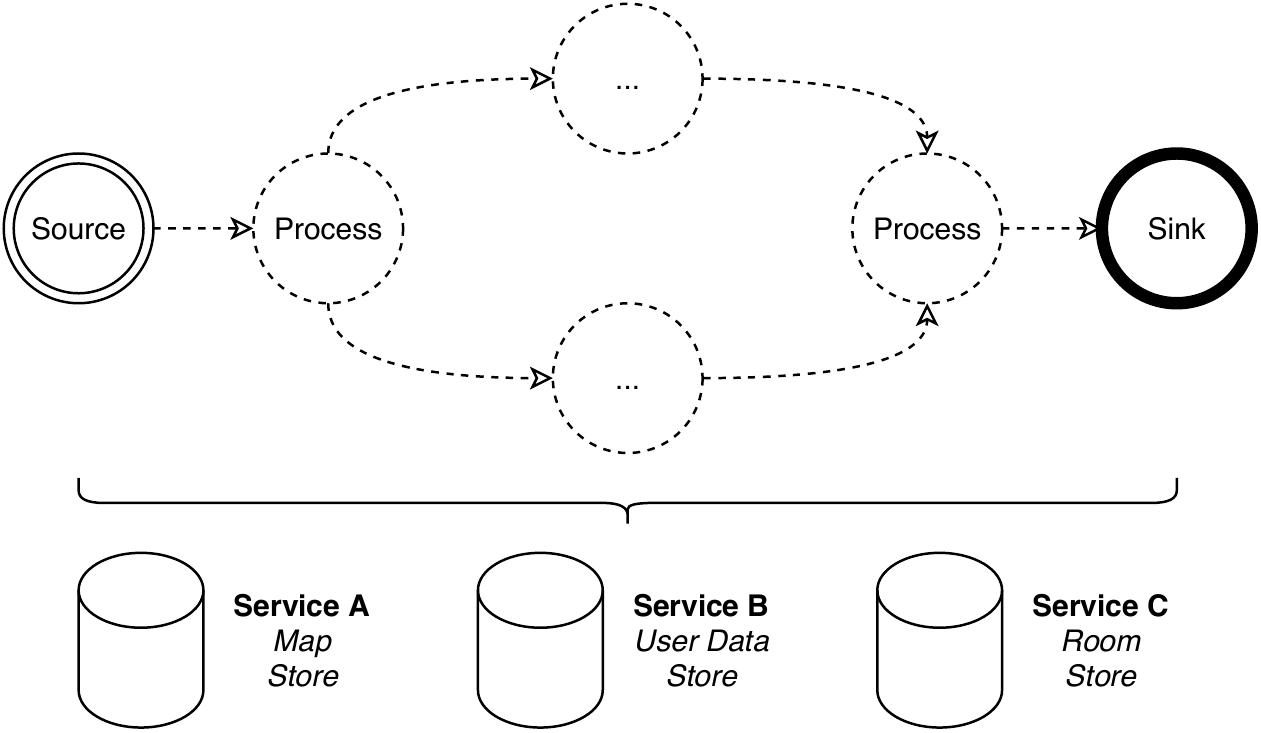}
	\Description{Image illustrates the positioning model used in OpenHPS. A source and sink is shown with processing nodes inbetween. Underneath the nodes three services (represented as a data store) are shown for map storage, user data storage and room storage}
	\caption{Example OpenHPS positioning model}
	\label{fig:architecture}
\end{figure}

Extensions of these nodes, allowing for specific data flow shapes and common position processing nodes, are provided in our core component. Figure~\ref{fig:architecture} shows an example of a \emph{positioning model} that has a source node, a sink node and four processing nodes connected in a graph structure. This positioning model describes a configured computational model aimed for processing sensor and positioning data~\cite{lee1995dataflow}. Similar to existing streaming or pipelining frameworks, the graph can contain data flow shapes that manipulate the flow of data frames. Examples of such shapes include, but are not limited to balance nodes, data frame chunking, debouncing and merging of data objects and their processed positions.

All nodes in a positioning model have access to a set of services that allow the storage of objects. In the given example, three services are added for the storage of map, user data and room information. In our implementation, sink nodes always store data objects contained in received data frames. However, every node has the ability to fetch or insert new data into available services. This persistence allows for the storage of landmark objects, similar to the JSR-179~specification~\cite{di2005indoor}. At the same time, these services can be used as an interface to fetch the latest position without requiring a specific implementation in the sink.

The positioning model can be created by using a builder pattern as illustrated in Listing~\ref{lst:openhps-model}. This builder creates the immutable properties of the model, including data services and the flow of data from source to sink. Models can have multiple flow shapes, each with one or more sources, processing nodes and sinks.

\begin{listing}[htb]
	\begin{openhps}
ModelBuilder.create()
	.addService(/* ... */)
	.addShape(GraphBuilder.create()
		.from(/* ... */)
		.via(/* ... */)
		.to(/* ... */))
	.build().then((model: Model) => { /* ... */ });
	\end{openhps}
	\vspace{-0.2cm}
	\caption{Creation of a positioning model}
	\label{lst:openhps-model}
\end{listing}

\begin{figure}[htb] 
	\centering
	\includegraphics[width=\columnwidth]{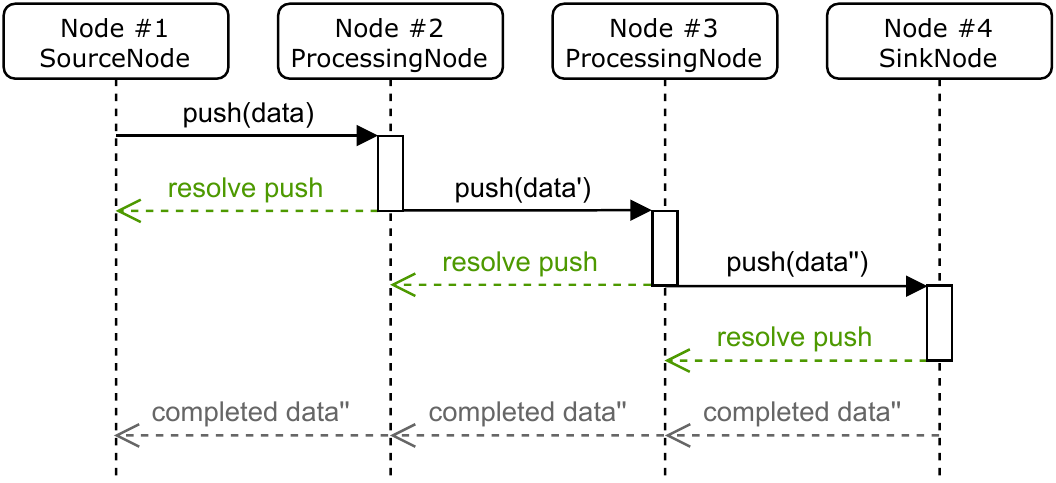}
	\Description{Swimlane consisting of four nodes. The image shows how a push is resolved after this push is processed by a node. The sink has an event that is sent to all other nodes to indicate the the processing is completed.}
	\vspace{-0.5cm}
	\caption{Data being pushed by a source}
	\label{fig:node-push}
\end{figure}

In Figure~\ref{fig:node-push}, data is being pushed by an \emph{active source} node. Processing nodes will process the data and push the modified frame to their output nodes. Push and pull actions are promise-based and resolved whenever the node finishes processing the frame. This allows for non-blocking asynchronous requests. The resolved push promise (indicated in green) gives an indication that the processing of the push is finished. However, it does not provide knowledge on whether or not the frame is processed by the complete network. To indicate this, sink nodes that receive a frame will emit a \mintinline{typescript}{completed} event that includes the data frame identifier and list of persisted object identifiers.

\begin{figure}[htb] 
	\centering
	\includegraphics[width=\columnwidth]{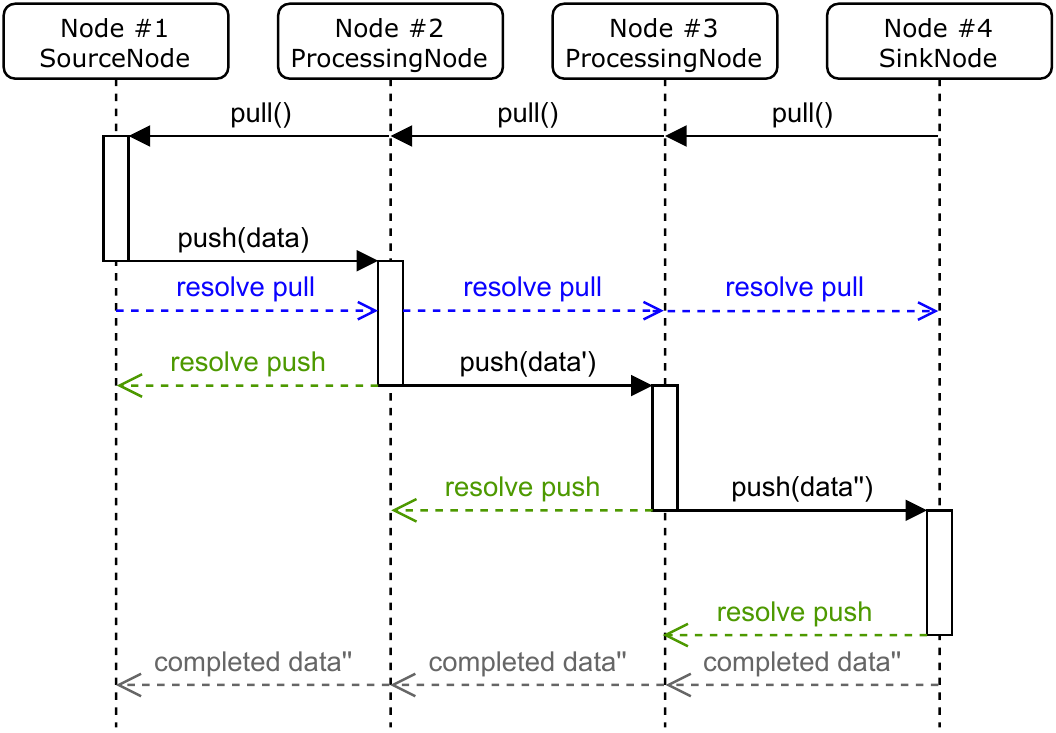}
	\Description{Swimlane consisting of four nodes. The image shows how a pull request initated by the sink is forwaded by each node until it reaches the source. The source sends a response as a push (see text) and resolves the pull request. Meanwhile, the push is processed and resolved in a similar manner as the previous figure.}
	\caption{Data being pushed by source after receiving a pull}
	\label{fig:node-pull}
\end{figure}

With the swim lane shown in Figure~\ref{fig:node-pull}, the data is not automatically pushed by the source node. A downstream node such as a sink will send a \mintinline{typescript}{pull()} request to its input nodes. If these nodes cannot provide a frame of their own, the \mintinline{typescript}{pull()} request is forwarded to their respective input nodes. If the source has data available, a response to this pull is provided asynchronously. As mentioned in the beginning of this section, a \mintinline{typescript}{pull()} response will use the same invocation as a \mintinline{typescript}{push()}. In that case, the pull promise is resolved right after the source sends this push as indicated by the blue resolve chain in Figure~\ref{fig:node-pull}.

\begin{figure}[htb] 
	\centering
	\includegraphics[width=\columnwidth]{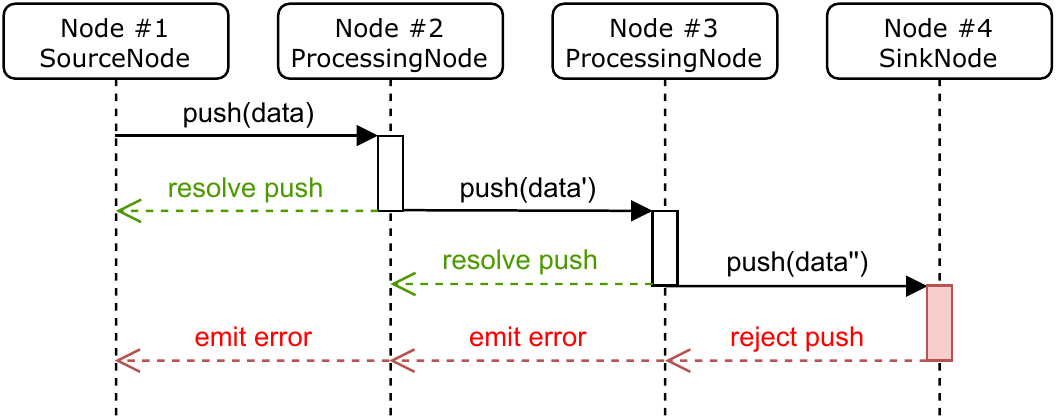}
	\Description{Swimlane consisting of four nodes. The image shows how an error is handled. Push promises are resolved when processed successfully. In the figure an error is thrown at the sink that does not resolve the promise, but rejects it. The node that initiated this rejected push will emit an error upstream.}
	\vspace{-0.5cm}	
	\caption{Error handling in \mintinline{typescript}{push()} request}
	\label{fig:node-push-error}
\end{figure}

As promises are resolved after the data frame is processed by a node, upstream nodes in the process chain cannot determine whether data has been processed successfully. Figure~\ref{fig:node-push-error} shows a \mintinline{typescript}{push()} request that throws an error at the sink node (e.g.~failure to store). An error event is triggered on previous node(s). By default, these nodes will chain the error to upstream nodes. However, each node can act upon this error in its individual implementation.

Nodes are implemented by developers on a high level of abstraction compared to other stream processing frameworks. Developers do not have the ability to push or pull from specific incoming or outgoing edges. Listing~\ref{lst:sourcenode} shows two custom source nodes. The pull-based source node on lines 1~to~7 implements the \mintinline{typescript}{onPull()} function that is called whenever the source receives a \mintinline{typescript}{pull()} request. This function expects a promise of a data frame. Internally, the extended source node class will push this data frame as shown in Figure~\ref{fig:node-pull}. With the push-based source (lines 9~to~22), the \mintinline{typescript}{onPull()} is unused. Instead, a timer is created that pushes a new data frame every 1000~milliseconds. A similar abstraction exists for sink nodes with the \mintinline{typescript}{onPush()} function.

\begin{listing}[htb]
	\begin{openhps}
export class PullBasedSource extends SourceNode<DataFrame> {
	public onPull(): Promise<DataFrame> {
		return new Promise((resolve) => {
			resolve(new DataFrame(this.source));
		});
	}	
}

export class PushBasedSource extends SourceNode<DataFrame> {
	constructor(source: DataObject) {
		super(source);
		this.on('build', () => {
			setInterval(this._generate.bind(this), 1000);
		});
	}
	private _generate(): void {
		this.push(new DataFrame(this.source))
	}
	public onPull(): Promise<DataFrame> { 
		return Promise.resolve(undefined); 
	}
}
	\end{openhps}
	\vspace{-0.2cm}
	\caption{Push- and pull-based SourceNode classes}
	\label{lst:sourcenode}
\end{listing}

Similar to sources and sinks, processing nodes are abstracted. Any \mintinline{typescript}{pull()} requests to these nodes are automatically forwarded to the incoming nodes, as these process nodes do not generate new data frames. Developers are expected to implement a \mintinline{typescript}{process()} function manipulating a frame or individual objects within a frame.

\begin{figure*}[t] 
	\centering
	\includegraphics[width=\textwidth]{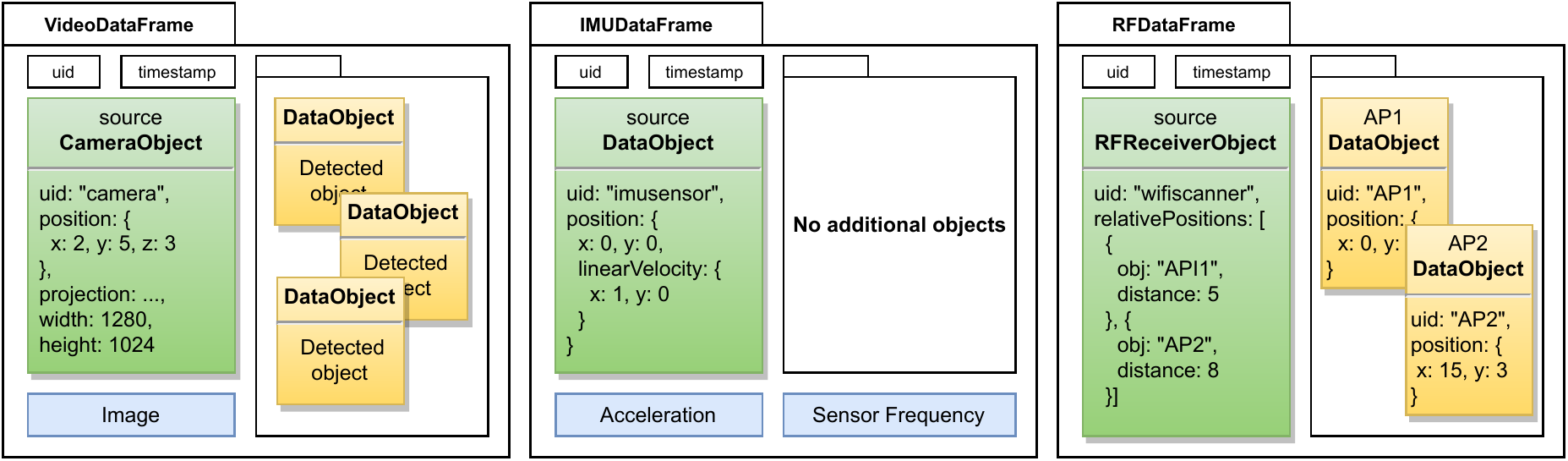}
	\Description{Three data frames represented as packages are depicted. Each frame has a separate block for the source of the data frame, an identifier and timestamp. An internal package in each data frame indicates the collection of objects that this data frame can contain. For the video data frame, we include an image attribute inside the data frame package. Three data objects with the caption ``detected object'' are shown in the video data frame. For the IMU data frame, acceleration and sensor frequency attributes are added to the frame. This frame does not have any additional objects. Finally, the third example is an RF data frame that contains a source object (receiver) with two relative positions to other objects (AP1 and AP2). These two objects are included in the additional objects of this RF data frame.}
	\vspace{-0.2cm}
	\caption{Data frame content examples}
	\label{fig:dataframe-examples}
\end{figure*}

\subsubsection{Data Frame} \label{subsubsec:data-frame}
Data that is pushed through the positioning model is represented within data frames, generated by a source node. This ensures that the origin of data can be determined via some collection of metadata. The data contained in these frames includes (but is not limited to) the following attributes:

\begin{itemize}
	\item \textbf{Unique identifier}: Each frame generated by a source is uniquely identified. This ensures that frames which are being processed by multiple processing nodes in parallel can be merged at a later stage in the stream.
	\item \textbf{Timestamp}: Required for determining when the data was created or obtained. When working with multiple sources that capture data of the same tracked actor, the timestamps will be used to merge the data frames. A timestamp is kept for the creation of each data frame by the source. This timestamp can also be used for time-based calculations such as applying velocity to a position. Using this timestamp instead of the system time results in a more deterministic output.
	\item \textbf{Source data object}: This is the data object that obtained the sensory data (e.g.~the camera object or RF~receiver). It is not always the object that is being tracked, but it can be required in order to determine the position of other objects (see actors in Section~\ref{subsec:actors}). Similar to the timestamp and identifier, the source data object can be used to specify certain criteria on how data frames or positions should be merged. 
	\item \textbf{Data objects}: Data objects include everything that is of relevance to the positioning (e.g.~the tracking and tracked actor). This also includes reference spaces needed for the positioning as pointed out later in Section~\ref{subsubsec:space}. By grouping the data objects in the same data frame, nodes do not have to access any services to get this relevant information.
\end{itemize}

In order to demonstrate the content of data frames, Figure~\ref{fig:dataframe-examples} depicts three situations where data is contained in frames. The first example shows a data frame created by a camera source. This camera object has a certain position and projection matrix. Linked to the data frame is a single image (i.e.~video frame) captured by this source. During the processing of the image, objects can be detected and added to this frame before being pushed further downstream. In the second example we show data obtained by an accelerometer. The source object has a velocity and position, the frame itself contains the current acceleration and sensor frequency. This information can be used by a processing node to add the acceleration to the existing velocity. In our third and final example, we show a data frame created by a Wi-Fi scanner. The scanner (source) has two relative distances to access points~(AP). The information, mainly the position of these access points is included in the frame.

\subsubsection{Position} \label{subsubsec:position}
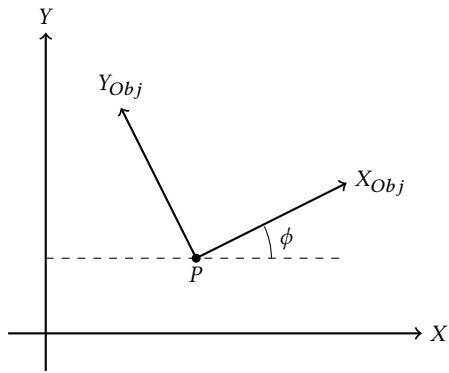
\begin{figure}[htb]
\begin{tikzpicture}
\draw[->,thick] (-0.5,0)--(5,0) node[right]{$X$};
\draw[->,thick] (0,-0.5)--(0,4) node[above]{$Y$};
\draw[->,thick] (2,1)--(4,2) node[right]{$X_{Obj}$};
\draw[->,thick] (2,1)--(1,3) node[above]{$Y_{Obj}$};
\draw[-,dashed] (0,1)--(4,1);
\draw[-] (3,1) arc (0:25:1) node[at start, above right]{$\phi$};
\node (P) at (2,1) [fill,circle,inner sep=1.2pt] {};
\draw (P) node [below] {$P$};
\end{tikzpicture}
\Description{The figure shows a large X and Y axis. In the first quadrant, two other axis are shown at a slight angle (phi).}
\captionof{figure}{Position representation} \label{fig:position-representation}
\end{figure}

Similar to existing work~\cite{gu2009survey,liu2007survey}, OpenHPS distinguishes between \emph{relative} and \emph{absolute} positions. Absolute positions represent a fixed position in a specified \emph{space} while relative positions indicate the position relative to another object. Absolute positions contain the following information:

\begin{itemize}
	\item \textbf{Timestamp}: The time when the absolute position has been recorded or modified. The timestamp can be set by the sensor or by a processing node that calculated the position.
	\item \textbf{Accuracy}: General position accuracy with the same unit as the position itself. In the context of a hybrid positioning system, the accuracy can be used as a weight when merging with other calculated positions.
	\item \textbf{Orientation}: Stationary orientation of the data object at the recorded position. This orientation is relative to the $X$-axis and is represented in quaternions. However, it is possible to convert the quaternion representation to (and from) Euler or axis angles. 
	\item \textbf{Linear velocity}: Linear velocity at the recorded position, relative to the orientation of the object (see Figure~\ref{fig:position-representation}) using the axis~$X_{Obj}$~and~$Y_{Obj}$ of the point~$P$ with orientation~$\phi$.
	\item \textbf{Angular velocity}: Similar to linear velocity, the angular velocity is relative to the orientation of the object.
	\item \textbf{Position vector}: Each position can be converted to a three-dimensional vector, which enables the use of 2D~positions in 3D~reference spaces.
	\item \textbf{Unit}: Length unit of the position. This unit applies to the position vector and its accuracy.
\end{itemize}

\noindent Relative positions have the following attributes:

\begin{itemize}
	\item \textbf{Timestamp}: Similar to the absolute position, this is the time when the relative position has been recorded.
	\item \textbf{Accuracy}: General position accuracy in a specified unit. 
	\item \textbf{Reference object}: The referenced data object to which the position is relative to.
	\item \textbf{Reference value}: The value that determines the relative position to the reference object. This can be a distance, angle or velocity.
\end{itemize}

By default, OpenHPS and its core positioning algorithms support 2D, 3D and geographical coordinates. Developers can further extend the coordinate space with higher-level absolute and relative positions. Positions can be stored with a specified unit (i.e.~length unit for absolute positions) in order to offer developers flexibility in the stored precision. Linear and angular velocity values are converted to a fixed unit ($m/s$ for linear and $rad/s$ for angular velocity). However, this can be customised by extending the velocity objects.

The \emph{position} terminology is used throughout the API as opposed to \emph{location} or \emph{pose}. Pose is a term that is often used when defining a position and orientation in a three-dimensional space. However, with the support of 2D~positions, this term was not favourable. Location is described by the English Oxford Dictionary as \emph{``a particular place or position''}. This abstraction of ``place'' led us to our final decision of choosing the more precise \emph{position} terminology.

\subsubsection{Data Object} \label{subsubsec:data-object}
A data object represents anything that is relevant to the positioning. It can be the tracked object, the tracking object or a landmark needed for the relative positioning. Each object contains the following attributes:

\begin{itemize}
	\item \textbf{Unique identifier}: Data objects are uniquely identified, either by a supplied identifier or a random~UID. Optionally, a developer can provide a more user-friendly display name.
	\item \textbf{Absolute position}: Data objects store their last known absolute position. The stored position is always relative to the global reference space introduced later in Section~\ref{subsubsec:space}. The relevance of this last known position can be determined using its timestamp and developers can request the transformed position in their own reference space.
	\item \textbf{Relative positions}: These are relative positions to other reference objects. Each object can have multiple types of positions relative to different objects. This allows a data object to have a relative distance, angle and velocity to the same object. 
	\item \textbf{Parent object}: A data object can specify its parent. This can be useful for indicating that individual sensor objects belong to the same tracked actor.
\end{itemize}

Depending on what the data object represents, it can be extended to store the information needed for its representation. In Listing~\ref{lst:openhps-dataobject}, we create a basic data object of a user who is uniquely identified by their e-mail. During the creation of this object, we set the current position to a geographical coordinate. 

\begin{listing}[htb]
	\begin{openhps}
const object = new DataObject("mvdewync@vub.be");
object.displayName = "Maxim Van de Wynckel";
object.setPosition(new GeographicalPosition(50.82075, 4.39234));
	\end{openhps} 
	\vspace{-0.2cm}
	\caption{Creation of a \mintinline{typescript}{DataObject}}
	\label{lst:openhps-dataobject}
\end{listing}

Data objects can be created and modified without those changes being persisted in the positioning model. In order to detect persisted changes, a listener can be added to the data object service as shown in Listing~\ref{lst:openhps-dataobject-listener}. 

\begin{listing}[htb]
	\begin{openhps}
const service = myModel.findDataSerice(DataObject);
service.on('insert', (uid, changedObject) => {
	if (uid === object.uid)
		console.log(changedObject.getPosition());
});
	\end{openhps}
	\vspace{-0.2cm}
	\caption{Listener for data object changes}
	\label{lst:openhps-dataobject-listener}
\end{listing}

\subsubsection{Reference Space} \label{subsubsec:space}
Reference spaces are data objects that represent spaces which are used for absolute positions. Using these reference spaces, absolute positions created in a different space can easily be identified and transformed to the global reference space created when building a model.

\begin{listing}[htb]
	\begin{openhps}
		const refSpace = new ReferenceSpace(model.referenceSpace)
		.unit(LengthUnit.CENTIMETER)
		.translation(10, 10, 0)
		.scale(1, 1, 0)
		.rotation(0, 0, 0, AngleUnit.RADIANS);
	\end{openhps}
	\vspace{-0.2cm}
	\caption{Creation of a \mintinline{typescript}{ReferenceSpace}}
	\label{lst:openhps-referencespace-create}
\end{listing}

Listing~\ref{lst:openhps-referencespace-create} shows the creation of a reference space relative to the global space represented by \mintinline{typescript}{model.referenceSpace}. This reference space has an origin offset. Absolute positions set when providing this reference space will automatically transform to the origin of the global space.

\noindent A reference space can transform the position, velocity and orientation in the following ways:

\begin{itemize}
	\item \textbf{Translation}: Translate the position with an origin offset. 
	\item \textbf{Rotation}: Rotate the position, orientation and angular velocity.
	\item \textbf{Scale}: Scale the position and linear velocity.
	\item \textbf{Perspective}: Transform the (inverse) perspective of the position (e.g.~the perspective of a camera).
	\item \textbf{Unit conversion}: Convert the unit of a position to a reference unit.
\end{itemize}

\noindent Reference spaces can be created to model different scenarios:

\begin{itemize}
	\item \textbf{Third-party positioning systems}: Frameworks like the  \mbox{WebXR~\cite{jones2019webxr}}~API manage their own origin and orientation based on the underlying hardware. The output of such third-party frameworks are high-level positions that should be aligned with the other positioning methods.
	\item \textbf{Sensor placement}: Developers can model a reference space for sensors that have a static offset or rotation (e.g.~a motion sensor that is placed upside down).
	\item \textbf{Calibrated reference space}: Some sensors require a calibration (either automatic or by manual user input). A goal of OpenHPS is to easily persist this type of calibration.
	\item \textbf{Map storage}: As a data object, a reference space can be extended to store environment map information as outlined in our functional requirements.
\end{itemize}

In Listing~\ref{lst:openhps-referencespace-setcurrentposition}, we set the current position of a data object to $(5, 5, 5)$ using the previously created reference space shown in Listing~\ref{lst:openhps-referencespace-create}. Internally, the stored position of \mintinline{typescript}{myObject} will be the transformed position with coordinates $(-5, -5, 5)$.

\begin{listing}[htb]
	\begin{openhps}
myObject.setPosition(new Absolute3DPosition(5, 5, 5), refSpace);
	\end{openhps}
	\vspace{-0.2cm}
	\caption{Setting the object position in a reference space}
	\label{lst:openhps-referencespace-setcurrentposition}
\end{listing}

As these spaces are data objects, they are uniquely identified and can have a parent object or space. This parent allows for abstract reference spaces such as \emph{rooms}, \emph{floors} and \emph{buildings}. These types of abstractions allow us to use different positioning methods per floor that are stored in a global reference space representing a building.

\subsubsection{Services} \label{subsubsec:service-design}
Each positioning model can have multiple services. A service can be accessed by all nodes in that model to perform certain general actions ranging from communication services that handle the data between remote nodes, to data services that store data frames, objects or other relevant information.

A data service serialises and stores information. By default, our core~API offers data services for:

\begin{itemize}
	\item \textbf{Data objects}: To store the processed objects and their last known position. This can also be used as a persistent storage for landmarks used in the positioning.
	\item \textbf{Node data}: Node-specific data about \mintinline{typescript}{DataObject}s can be stored. This can be useful for intermediate calculations by noise filtering algorithms or sensor fusion techniques.
	\item \textbf{Trajectories}: Historical position data of \mintinline{typescript}{DataObject}s. Drivers can be implemented for storing this information in specialised databases such as MobilityDB~\cite{zimanyi2019mobilitydb}.
\end{itemize}

Normal services in our framework include, but are not limited to a \emph{time service} that allows developers to synchronise the time between multiple machines, and a \emph{worker service} that acts as a (remote) proxy for data services.

Listing~\ref{lst:service-code} shows examples of how a service can be retrieved from the model. Nodes can retrieve a data service by providing either the class of an object, an object instance or the class name of the object. This allows the use of difference services for different types of \mintinline{typescript}{DataObject}s.

\begin{listing}[htb]
	\begin{openhps}
// Finding a data service by class
this.model.findDataService(DataObject);
// Finding a data service object instance
this.model.findDataService(myObject);
// Finding a data service by name
this.model.findDataService("RFDataObject");
	\end{openhps}
	\vspace{-0.2cm}
	\caption{Retrieving a data service from a model}
	\label{lst:service-code}
\end{listing}

\subsubsection{Measurement Units} \label{subsubsec:units}
Unlike many positioning frameworks aiming for geographical positioning, OpenHPS aims to support a wide range of use cases ranging from small scale to celestial positioning. We provide a unit system consisting of the \mintinline{typescript}{Unit} and \mintinline{typescript}{DerivedUnit} objects. A derived unit consists of multiple units with a specific power and offset. Math.js~\cite{jong2014math} offers a similar unit system with the possibility to automatically evaluate and convert units. While this allows for the easy creation of derived units, it is not necessary for our framework.

\begin{listing}[htb]
	\begin{openhps}
// Time unit called 'second'
const second = new TimeUnit('second', {
	// Unit for 'time'
	baseName: 'time',
	// Also called 's', 'sec' or plural
	aliases: ['s', 'sec', 'seconds'],
	// Supports decimal prefixes (milli, micro, ...)
	prefixes: 'decimal',
});

// Millisecond is a second with the prefix specifier milli
const millisecond = second.specifier(UnitPrefix.MILLI);

const minute = new TimeUnit('minute', {
	baseName: 'time',
	aliases: ['m', 'min', 'minutes'],
	// Minute can be defined as 60 * 1 second
	definitions: [{ magnitude: 60, unit: 's' }],
});
	\end{openhps}
	\vspace{-0.2cm}
	\caption{Unit creation}
	\label{lst:code-baseunit}
\end{listing}

Listing~\ref{lst:code-baseunit} shows the creation of a base unit \texttt{second} for time. During its creation the developer can specify aliases for the unit and similar to Math.js, a unit can have a set of unit prefixes. This allows the use of ``millisecond, microsecond, nanosecond, \ldots{}'' without specifically creating individual units for these specifiers. Note that aliases can be provided to optionally allow the units to be converted to string evaluators of other mathematical modules.

When creating a new unit, the developer should specify the base unit. For the minute example in Listing~\ref{lst:code-baseunit} this is done by creating a definition for converting minutes to seconds (using a magnitude of 60 for the unit seconds).

\begin{listing}[htb]
	\begin{openhps}
const radSecond = new DerivedUnit('radian per second', {
		baseName: 'angularvelocity',
		aliases: ['rad/s', 'radians per second'],
	})
	.addUnit(AngleUnit.RADIAN, 1)
	.addUnit(TimeUnit.SECOND, -1);

const degreeSecond = radSecond.swap(
	[AngleUnit.DEGREE],
	{
		baseName: 'angularvelocity',
		name: 'degree per second',
		aliases: ['deg/s', 'degrees per second'],
	});

const degreeMinute = radSecond.swap(
	[AngleUnit.DEGREE, TimeUnit.MINUTE],
	{
		baseName: 'angularvelocity',
		name: 'degree per minute',
		aliases: ['deg/min', 'degrees per minute'],
	});
	\end{openhps}
\vspace{-0.2cm}
\caption{Derived unit creation}
\label{lst:code-derivedunit}
\end{listing}

In order to use a unit that is derived from other base units, a \mintinline{typescript}{DerivedUnit} can be created as shown in Listing~\ref{lst:code-derivedunit}. The developer provides a name of the unit and adds the units that are contained in the derived unit (lines 5~and~6) along with their magnitude. Variants on derived units can be created by \emph{swapping} a unit (lines 9~and~17).

\subsection{Framework Implementation} \label{subsec:implementation}
OpenHPS is implemented in TypeScript\footnote{\href{https://www.typescriptlang.org}{https://www.typescriptlang.org}}, a type-safe superset of JavaScript. It can be executed as a client-side browser application, hybrid mobile applications, on JavaScript supported embedded systems such as Espruino and even as a server-side application using Node.js\footnote{\href{https://nodejs.org/en/about/}{https://nodejs.org/en/about/}} or Deno\footnote{\href{https://deno.land}{https://deno.land}}.

The ability to run our positioning model on a large range of server and client devices enables the decentralisation mentioned in the functional requirements. Additional remote components such as the socket~API outlined in Section~\ref{subsubsec:modularity} allow for other programming languages to be supported as well.

\subsubsection{Serialisation}
Data frames and contained objects are serialisable throughout the framework. This functionality is implemented using an extension of TypedJSON\footnote{\href{https://github.com/JohnWeisz/TypedJSON}{https://github.com/JohnWeisz/TypedJSON}} that adds the ability for polymorphic data types. The detection of such data types is necessary for allowing developers to create additional position or data objects without having to recreate all classes where these are used.

\begin{listing}[htb]
	\begin{openhps}
{
	"createdTimestamp":1606501972983302,
	"uid":"8865727c-7c98-4a8d-a33c-506d2650e59d",
	"position":{
		"x":-4.07093248547983,
		"y":55.59130128032057,
		"timestamp":1606502001594449,
		"velocity":{
			"linear":{
				"x":-0.27608249684331726,
				"y":0.3606549076013354,
				"z":0.013291033512841348
			},
			"angular":{
				"x":-3.9937982517329886,
				"y":0.2311694373502423,
				"z":-0.5070813464456928
			}
		},
		"orientation":{
			"x":-0.09754179767548248,
			"y":0.15388368786071302,
			"z":0.04266920115206052,
			"w":0.9823363719162936
		},
		"unit":{
			"name":"centimeter"
		},
		"referenceSpaceUID":"5582d63d-c7af-4624-9fed-6ce0d9036f62",
		"accuracyUnit":{
			"name":"meter"
		},
		"__type":"Absolute2DPosition"
	},
	"relativePositions":[],
	"__type":"DataObject"
}
	\end{openhps}
	\vspace{-0.2cm}
	\caption{Serialised \mintinline{typescript}{DataObject}}
	\label{lst:openhps-serialisation-dataobject}
\end{listing}

Listing~\ref{lst:openhps-serialisation-dataobject} shows a serialised data object created with sensor data retrieved from a \emph{Sphero Mini}\footnote{\href{https://sphero.com/products/sphero-mini}{https://sphero.com/products/sphero-mini}} toy. The main \mintinline{typescript}{DataObject} and \mintinline{typescript}{Absolute2DPosition} have a \mintinline{text}{__type} key that defines the object type. Definitions of a unit are not included in the serialisation and its complete name is used to indicate the unit. This means that a custom unit should be available in all processes that are required to deserialise the unit.

\subsubsection{Performance} \label{subsubsec:performance}
One of the non-functional requirements mentioned in Section~\ref{subsec:requirements} is the ability to perform real-time data processing. In order to achieve these performance requirements, parts of the processing network can be run in their own thread, web worker or process. This threading is made possible due to the serialisability of data frames and objects, which allows the transmission of frames from one process or thread to another.

\begin{listing}[htb]
	\begin{openhps}
ModelBuilder.create()
	.addService(/* ... */)
	.from(/* ... */)
	.via(new WorkerNode((builder: GraphShapeBuilder) => {
		const { TrilaterationNode } = require('@openhps/core');
		builder.via(new TrilaterationNode())
	}, {
		poolSize: 4
	}))
	.to(/* ... */)
	.build().then(model => { /* ... */ });
	\end{openhps}
	\vspace{-0.2cm}
	\caption{Threaded node creation}
	\label{lst:openhps-threading}
\end{listing}

Listing~\ref{lst:openhps-threading} shows the creation of a model with parts of the graph going through a \mintinline{typescript}{WorkerNode}. This threaded node is initialised with a model builder function evaluated on the threaded process. If no data services are (re)initialised in this function, the data services of the main thread are made accessible in the individual threads.

A \mintinline{typescript}{WorkerNode} can also run a larger portion of a process network that is declared in a separate file. This is more developer friendly than having to import all the nodes within a builder function. Listing~\ref{lst:openhps-threading-graph} shows the worker node named ``video'' being created in the main thread (lines 2~to~6). Internally, this node is a graph created in \mintinline{text}{video.ts}. Pull requests to this node (line 8) will be forwarded to a pool of four workers.

\begin{listing}[htb]
	\begin{openhps}
// main.ts //
ModelBuilder.create()
	.addNode(new WorkerNode("video.ts", {
		poolSize: 4,
		name: "video"
	}))
	.from("video")
	.via(new TimedPull(1, TimeUnit.MILLISECOND))
	.to(/* ... */)
	.build().then(model => { /* ... */ });

// video.ts //
export default GraphBuilder.create()
	.from(/* ... */)
	.via(/* ... */)
	.to();
	\end{openhps}
	\vspace{-0.2cm}
	\caption{Threaded graph creation}
	\label{lst:openhps-threading-graph}
\end{listing}

As a simple demonstration of our worker node, we created a processing node calculating 5000~prime numbers for every received frame. This test was conducted on an Intel i7-6700HQ laptop~CPU with 8 logical cores, running Node.js~14.10. These 5000~prime numbers can be calculated 237.03 times per second without the overhead of data frames, objects and services. The data frames that we push contain a source object, position and velocity to simulate the amount of data normally serialised and communicated between the main process and workers. However, the contained data does not affect the time it takes to compute the prime numbers.

Table~\ref{table:benchmark} shows the results of our benchmark with one worker assigned to each logical CPU core. Performance is measured in frames per second~(FPS) represented by the amount of computed data frames received by the sink of our model. For each worker we indicate the speed-up compared to the sequential implementation. The overhead shown with a single worker is due to the serialisation and deserialisation of data, an operation that is not required when pushing in a sequential network.

\begin{table}[htb]
	\begin{tabular}{c|c|c|c}
		\textbf{\#workers} & \textbf{FPS} & \textbf{Error} & \textbf{Speed-up} \\ \hline
		Sequential & 229.04 & $1.19\%$ & - \\
		1 & 200.74 & $0.67\%$ & 0.88 \\
		2 & 389.44 & $0.56\%$ & 1.70 \\
		3 & 512.42 & $0.92\%$ & 2.24 \\
		4 & 616.29 & $1.15\%$ & 2.69 \\
		5 & 671.00 & $0.59\%$ & 2.93 \\
		6 & 746.07 & $0.67\%$ & 3.26 \\
		7 & 801.32 & $0.90\%$ & 3.50 \\
		8 & 822.47 & $0.69\%$ & 3.59 \\
	\end{tabular}
	\vspace{0.2cm}
	\captionof{table}{\mintinline{typescript}{WorkerNode} benchmark} \label{table:benchmark}
	\vspace{-0.8cm}
\end{table}

\subsubsection{Precision}
Calculations within the framework are made using JavaScript number operations. Time-critical operations use a time service that returns the time in a specific unit. This allows developers to extend the framework with additional modules such as microtime\footnote{\href{https://www.npmjs.com/package/microtime}{https://www.npmjs.com/package/microtime}} for more precise calculations. In addition, Decimal.js\footnote{\href{https://www.npmjs.com/package/decimal}{https://www.npmjs.com/package/decimal}} could be used with an extended position class to provide more precise number operations.

\subsubsection{Modularity} \label{subsubsec:modularity}
OpenHPS provides a modular API that splits the functionality of positioning methods and algorithms in different npm\footnote{\href{https://www.npmjs.com}{https://www.npmjs.com}} \emph{modules}. Using this method, developers
can extend on our core~framework or other components. It also prevents them from having to depend on very large modules, reducing the overall dependency size.

\begin{figure}[htb]
	\centering
	\includegraphics[width=\columnwidth]{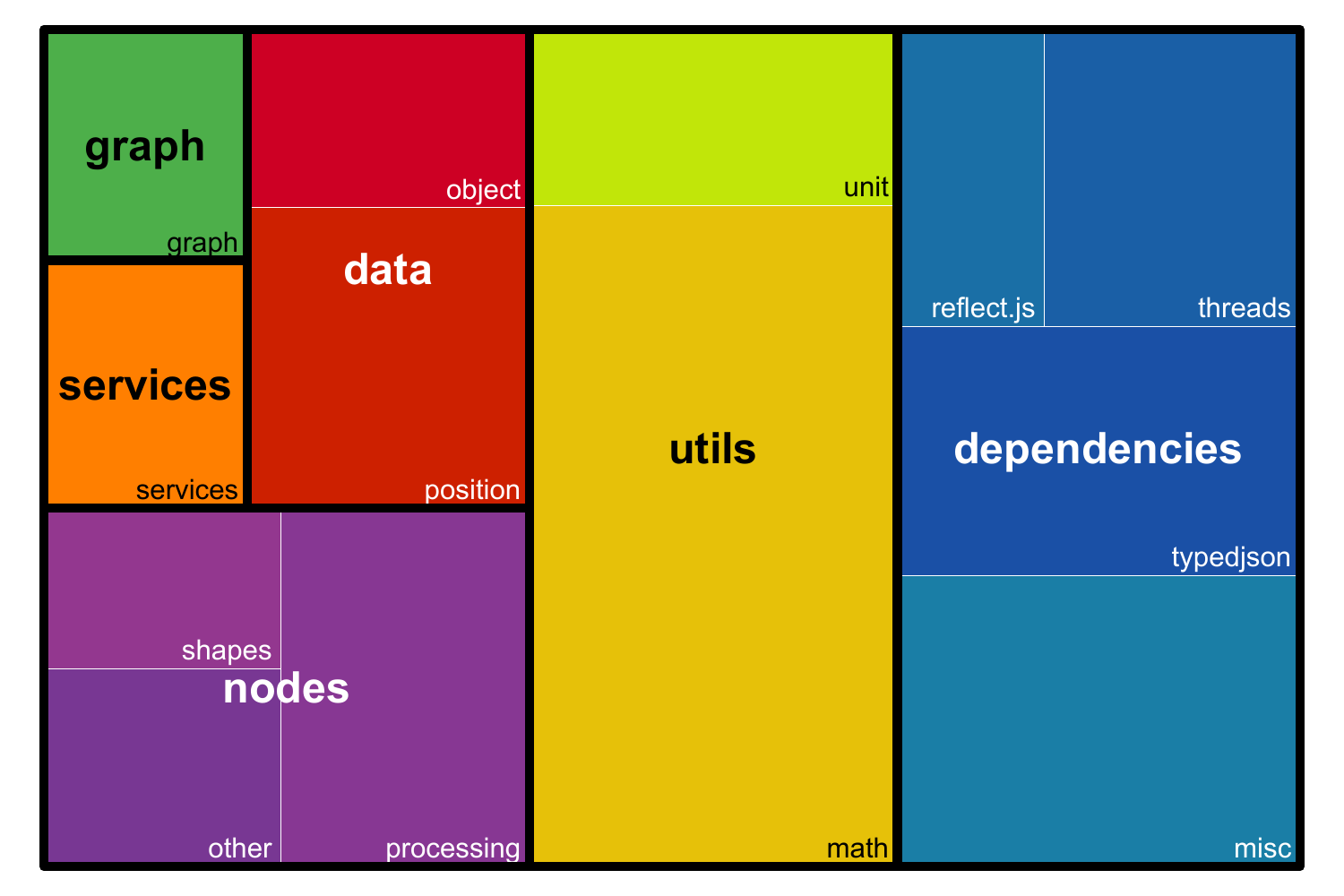}
	\Description{The figure shows a treemap visualisation of the content sizes. Noteworthy contents are mentioned in the text with aproximate percentages.}
	\vspace{-0.5cm}	
	\caption{@openhps/core minified web module tree map}
	\label{fig:openhps-size}
\end{figure}

Our core~API, named \emph{@openhps/core} is available for server and web deployment in the CommonJS~(CJS), ECMAScript~(ESM) and Universal Module Definition~(UMD) formats. Figure~\ref{fig:openhps-size} presents an overview of the core~components content size in its current state (version 0.2.0). Most of the file size is taken up by dependencies ($\approx\kern-0.5ex30\%$, indicated in blue) and the mathematical classes of Three.js\footnote{\href{https://threejs.org/docs/}{https://threejs.org/docs/}} ($\approx\kern-0.5ex22\%$, as part of the mathematical utilities in yellow). Default nodes such as processing nodes, graph shapes and common sink or source nodes account for $\approx\kern-0.5ex16\%$. The main purpose of the dependencies are to help with serialisation (i.e.~TypedJSON, Reflect.js). Mathematical classes such as quaternions, matrices and vectors from Three.js offer general operations for handling 2D and 3D position manipulation.

\noindent We list several examples of modules that can be used to extend the core functionality:

\begin{itemize}
	\item \textbf{Data storage}: By default, the core~API provides the possibility of using in-memory data storage. In order for this data to be persisted, additional components making use of different database management systems such as MongoDB, Redis or MobilityDB~\cite{zimanyi2019mobilitydb} can be applied.
	\item \textbf{Remote communication}: The remote APIs introduce a \mintinline{typescript}{RemoteNode} that can be added to the model. This node will transmit push (or pull) requests to a remotely connected model through either a REST~API, socket connection or a message broker such as MQTT~\cite{hunkeler2008mqtt}.
	\item \textbf{Positioning methods and algorithms}: The core~API offers basic processing nodes to determine a position (i.e.~trilateration, triangulation or fingerprinting) and can be extended with different components. Examples include techniques that require additional machine learning or computer vision libraries.
	\item \textbf{Symbolic positions}: Our core~API offers the 2D, 3D and geographical positioning. This can be abstracted to \emph{locations} or \emph{places} such as a room, building or site.
	\item \textbf{Third-party positioning systems}: Third-party positioning solutions can be integrated into OpenHPS by using modules that provide this interface.
\end{itemize}

\subsection{Demonstrator} \label{subsec:demonstrator}
Unlike some of the frameworks discussed in Section~\ref{sec:related-work} that are made to tackle a certain issue or goal, the core idea of OpenHPS is to combine different positioning concepts into one model.

As a non-trivial demonstrator, we provide a positioning system for a Sphero~Mini toy using the internal sensors and an external Logitech Brio\footnote{\href{https://www.logitech.com/en-us/product/brio}{https://www.logitech.com/en-us/product/brio}} camera. The Sphero provides raw sensor reading for the linear and angular velocity, raw accelerometer data, orientation and an internally computed position. This internal position is computed by the Sphero toy itself and makes use of the motor velocity, accelerometer and gyroscope.

We make use of the @openhps/core\footnote{\href{https://github.com/OpenHPS/openhps-core/}{https://github.com/OpenHPS/openhps-core/}}, @openhps/opencv\footnote{\href{https://github.com/OpenHPS/openhps-opencv/}{https://github.com/OpenHPS/openhps-opencv/}} and the use case-specific @openhps/sphero\footnote{\href{https://github.com/OpenHPS/openhps-sphero/}{https://github.com/OpenHPS/openhps-sphero/}} modules to construct a model that fuses these multiple sources together into one position. The model consists of four sources; the video input, internally computed position, the input that is sent to the Sphero and finally the dead reckoned position that is calculated by the framework itself using the provided velocity.

\begin{figure}[htb]
	\includegraphics[width=\columnwidth, keepaspectratio, page=1]{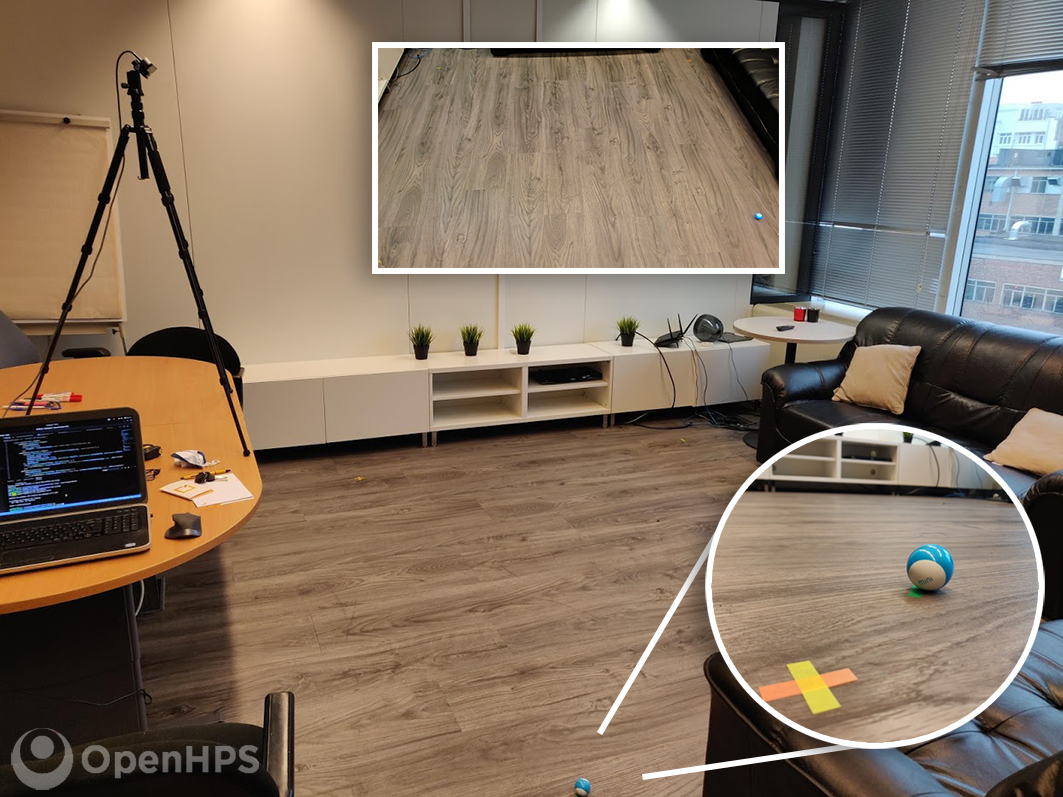}
	\Description{The figure shows a photo of our set up. A tripod is positioned high on top of a table, with the camera looking down at the ground. The figure contains an additional embedded photo of the view that the camera sees. The ground has an empty space with four yellow markers (not clearly visible on the photo). A zoomed in embedded image shows a blue ball next to a yellow marker.}
	\captionof{figure}{Demonstrator overview} \label{fig:demo-overview}
\end{figure}

Our setup is shown in Figure~\ref{fig:demo-overview}. We used yellow floor markers to define an area of $260\,cm \times 200\,cm$. The camera is positioned with a perspective view on the area and the start position of the Sphero is at the bottom right corner of the camera source.

For the scope of this demonstration, the Sphero performs a simple trajectory. The input for this device consists of an orientation (heading) and speed. Before
the start of our input trajectory, we manually calibrated the origin orientation using the provided mobile application for the Sphero. This provides us knowledge on the start orientation used by the internally calculated position which allowed us to define the reference spaces.

\begin{figure}[htb]
	\includegraphics[width=\columnwidth, keepaspectratio, page=1]{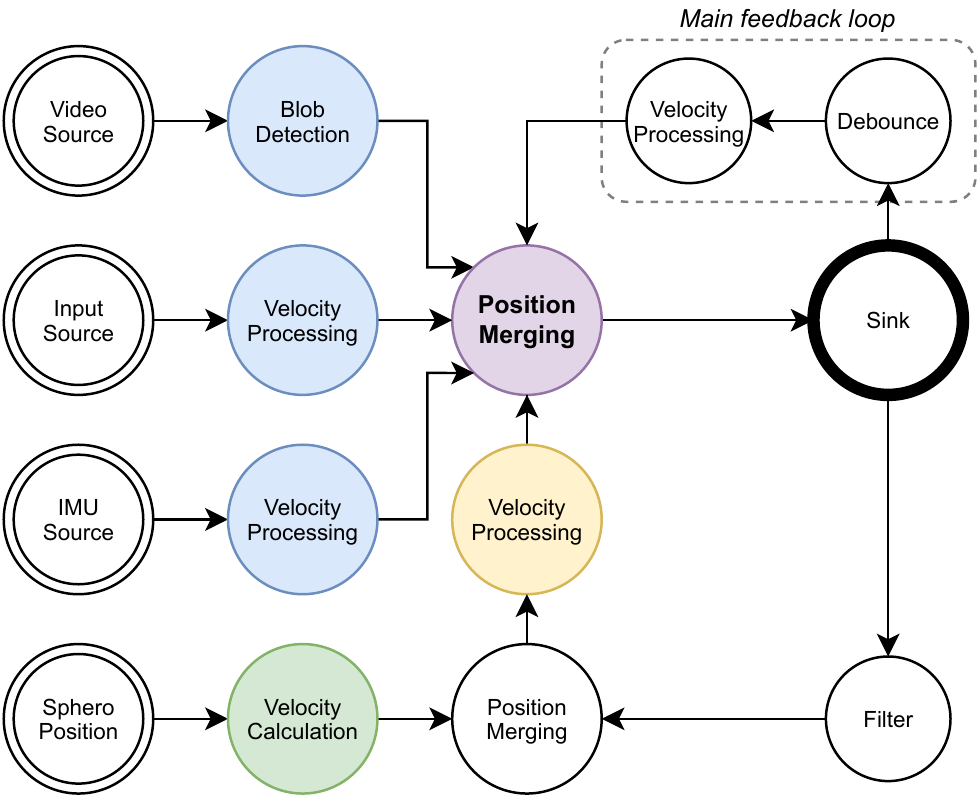}
	\Description{The figure shows a directed graph. Four vertices (nodes) are shown for the video source, input source, IMU source and Sphero position. The video source has directed link to a ``blob detection'' node. Both the input source and IMU source have a directed link to a ``velocity processing'' node. For the Sphero position, a directed link to a node with the caption ``velocity calculation'' is shown. This ``velocity calculation'' has a directed link to a ``position merging'' node that finally links to a ``velocity processing'' node. The blob detection and three remaining velocity processing nodes are linked to a single ``position merging'' node. This position merging node has a directed link to a sink. Finally, the sink has two outgoing links. The first outgoing link has the caption ``Main feedback loop'' and links to a debounce node that in its turn links to a velocity processing node and creates a loop to the single position merging node that was linked to by the blob detection and velocity processing nodes. The second outgoing link of the sink links to a filter that in its turn links to the position merging node used by the Sphero position source.}
	\captionof{figure}{Demonstrator positioning model} \label{fig:demo-model}
\end{figure}

Various methods exist to combine the aforementioned positioning methods. Figure~\ref{fig:demo-model} shows the simplified graph presentation of our demonstrated positioning model. Starting from the four different sources, we will discuss how each signal is processed and fused together. We use two feedback loops from our fused position to provide temporal information to our positioning model.

\begin{figure}[htb]
	\includegraphics[width=\columnwidth, keepaspectratio, page=1]{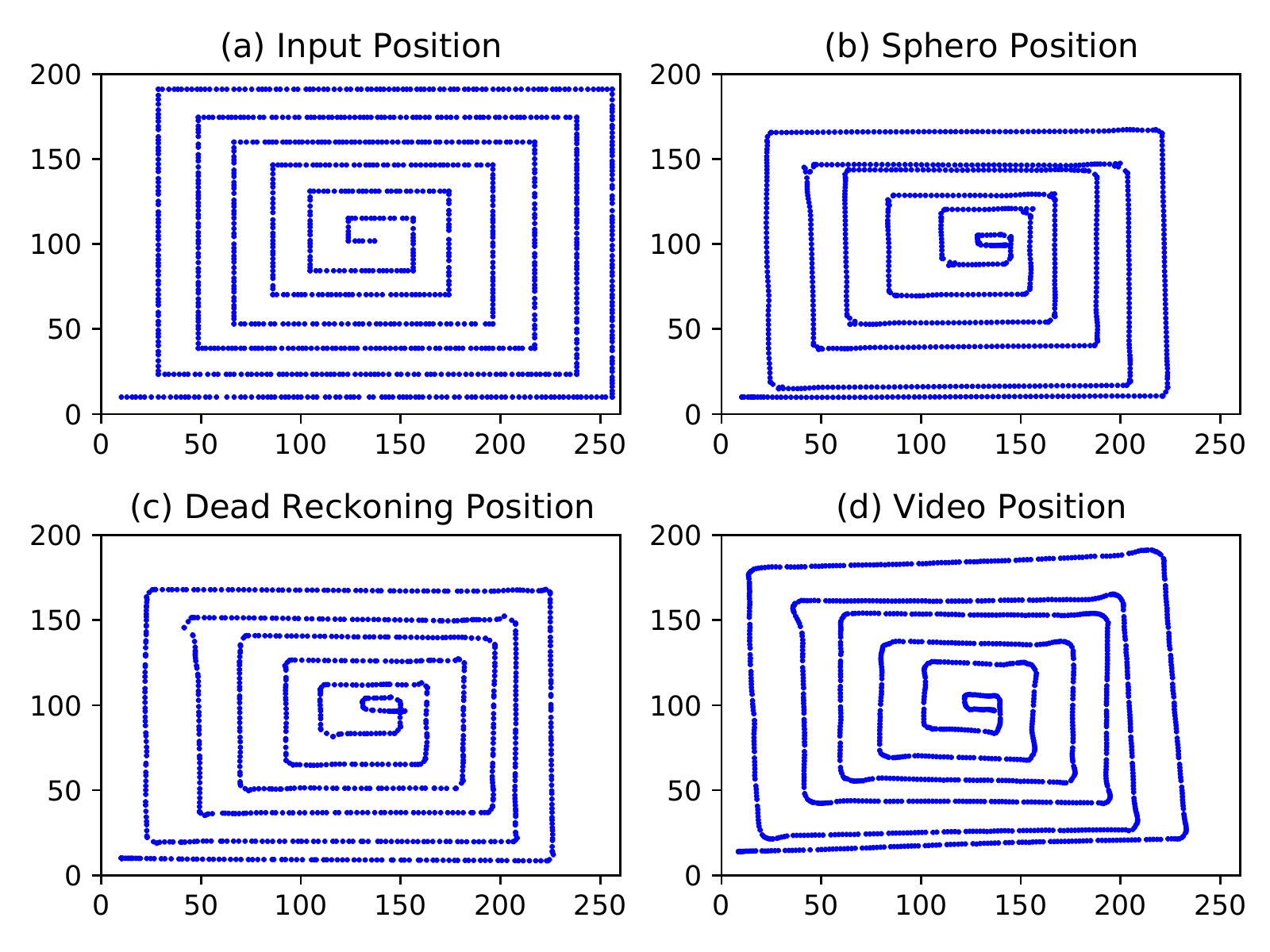}
	\Description{The figure shows the plots of the three individual positioning methods. The input position has very precise corners but has a larger X and Y axis movement than the other positioning methods shown in this figure. The Sphero position also has sharp corners, but the top most trajectories of the spiralling rectangle are very close together. The dead reckoning position determined in our framework does not have the issue of the lines being too close together, but has more rounded corners. Finally, the video position shows the spiralling rectangle. However, the first outer most spirals have a slight angle compared to other positioning methods.}
	\captionof{figure}{Individual position estimates for the given input~(a), including the internally calculated position~(b), dead reckoning position~(c) and video source~(d)} \label{fig:sensor-individual}
	\vspace{-0.2cm}
\end{figure}

The results of each independent source is shown in the trajectory scatter plots in Figure~\ref{fig:sensor-individual}. Each positioning method has a different frequency, resulting in a varying amount of data points used to determine the position. Our main feedback loop in Figure~\ref{fig:demo-model} ensures that the fused position never relies on a single source.

\subsubsection{Input Control}
Input to the Sphero is given using a heading (degrees), speed (0-255) and roll duration. We make the assumption that the Sphero~Mini has a maximum speed of $1\,m/s$ as documented on the product website\footnote{\href{https://support.sphero.com/article/6drb2qggx4-sphero-mini-faq}{https://support.sphero.com/article/6drb2qggx4-sphero-mini-faq}}.

As input trajectory, we provide a spiralling rectangle starting from an outer corner to the centre of the area with a speed of $150$ ($=0.58\,m/s$). We provide a basic roll duration of 4.2 seconds ($=2.436\,m$) for the X-axis and a roll duration of 3.2 seconds ($=1.856\,m$) for the Y-axis. Every turn, the duration of the movement along the X-axis is reduced by 168\,ms while the movement along the Y-axis is reduced by 128\,ms. This input is fed to our framework's velocity processing node resulting in the output shown in Figure~\ref{fig:sensor-individual}a.

\subsubsection{Visual Positioning}
The video source uses the OpenCV~\cite{opencv_library} library to capture a 30\,FPS camera feed from the Logitech Brio camera which has a perspective view of the floor. When processing the video stream, we create the inverse perspective view by manually specifying the position of four yellow markers on the floor. This creates a wrapped image rectangle of $1040\,px \times 800\,px$.

\begin{figure}[htb]
	\includegraphics[width=\columnwidth, keepaspectratio, page=1]{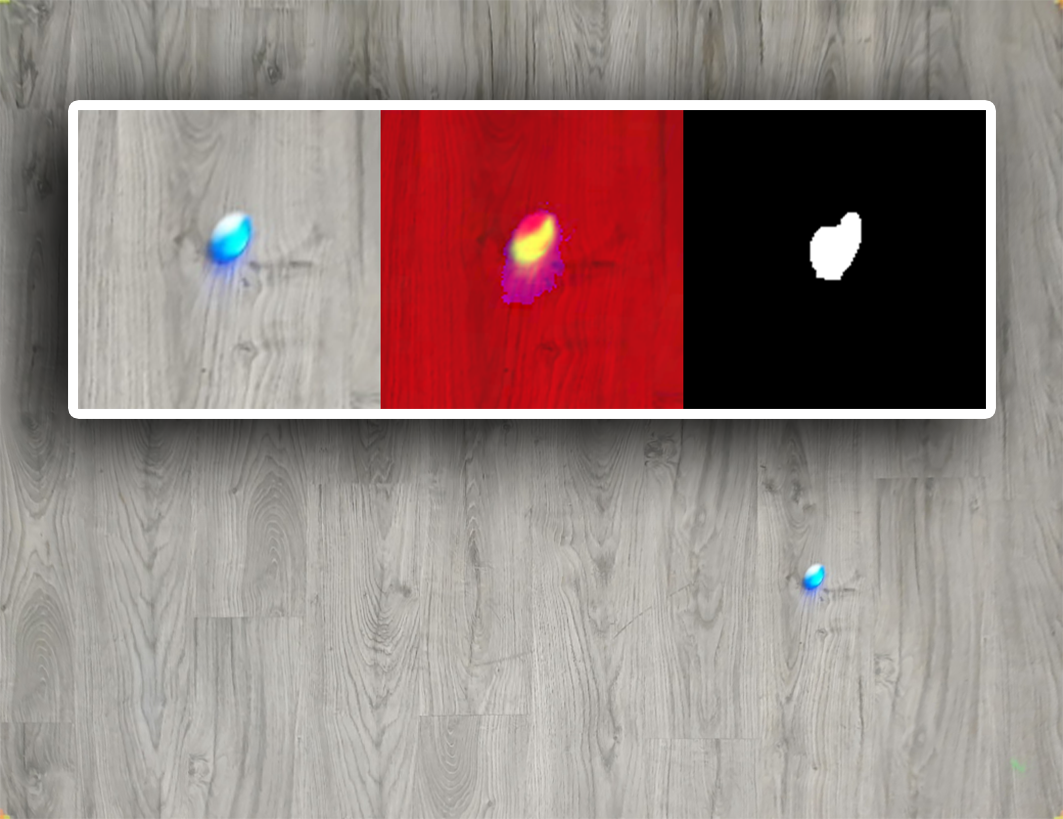}
	\Description{The figure shows a top down view of the ground where the tracking takes place. Three embedded images are shown that show the transformation from this top down view to the detected black and white blob.}
	\captionof{figure}{Conversion of wrapped video to blob} \label{fig:demo-wvs}
	\vspace{-0.2cm}
\end{figure}

Once the video source is wrapped, blob detection is used to determine the centroid position of the blue Sphero~Mini. We apply a colour mask that converts the image to an HSV colour space and performs a masking filter to only show the blue ball as illustrated in Figure~\ref{fig:demo-wvs}. Next, in Listing~\ref{lst:openhps-demo-visual-contour} we create a custom processing node that sets the position of our tracked object to the pixel position of the blob. As the accuracy for our position we take the square root of the blob area. A reference space is created (lines 1~to~4 in Listing~\ref{lst:openhps-demo-visual}) and applied to the output position (pixel coordinate) on line~25 to scale it to the corresponding rectangle dimensions.

Without interference from other sources, the video processing provides the output shown in Figure~\ref{fig:sensor-individual}d. We will use this source as our most accurate position, as it is the only available external positioning method.

\subsubsection{Internal Position} \label{subsubsec:internal-position}
In Figure~\ref{fig:sensor-individual}b we show the internal positioning calculated by the Sphero, converted to a certain reference space created with our calibrated orientation knowledge. Instead of using the raw position, we determine the displacement of this internal position (using the filtered feedback loop shown in Figure~\ref{fig:demo-model}) and apply this displacement to the fused position.

\begin{listing}[htb]
	\begin{openhps}
class ContourDetectionNode extends ProcessingNode<VideoFrame> {
	public process(frame: VideoFrame): Promise<VideoFrame> {
		return new Promise((resolve) => {
			let contours = frame.image.findContours(
				OpenCV.RETR_EXTERNAL, 
				OpenCV.CHAIN_APPROX_SIMPLE);
			if (contours.length >= 1) {
				// Sort contours by area
				contours = contours.sort((a, b) => a.area - b.area);
				// Select the contour with the largest area size
				const m = contours[0].moments();
				const center = new OpenCV.Vec2(
					m.m10 / m.m00, 
					m.m01 / m.m00);
				// Use the center as the 2D pixel position
				const position = new Absolute2DPosition(
					center.x, 
					center.y);
				position.unit = LengthUnit.CENTIMETER;
				position.accuracy = Math.sqrt(contours[0].area);
				frame.source.setPosition(position);
			}
			resolve(frame);
		});
	}
}
	\end{openhps}
	\vspace{-0.2cm}
	\caption{Contour detection processing node}
	\label{lst:openhps-demo-visual-contour}
	\vspace{-0.2cm}
\end{listing}

\begin{listing}[htb]
	\begin{openhps}
const videoSpace = new ReferenceSpace(defaultSpace)
	.translation(1040, 800)
	.rotation(new Euler(180, 180, 0, 'ZXY', AngleUnit.DEGREE))
	.scale(4, 4);
/* ... */
export default GraphBuilder.create()
	.from(new VideoSource(new CameraObject("sphero_video"), {
		autoPlay: true,
		fps: 30,
		// Do not fetch a frame if the webcam can not handle it
		throttleRead: true,
		source: new CameraObject("sphero_video")
	}).load("/dev/video2"))
	.via(new ImageTransformNode({
		src: [
			new OpenCV.Point2(307, 120),
			new OpenCV.Point2(1473, 87),
			new OpenCV.Point2(1899, 891),
			new OpenCV.Point2(20, 1024),
		],
		height: 800, // 200cm
		width: 1040  // 260cm
	}))
	.via(new ColorMaskProcessing({
		minRange: [90, 50, 50],
		maxRange: [140, 255, 255]
	}))
	.via(new ContourDetectionNode())
	.convertFromSpace(videoSpace)
	.to();
	\end{openhps}
	\vspace{-0.2cm}
	\caption{Graph shape video}
	\label{lst:openhps-demo-visual}
\end{listing}

\subsubsection{Dead Reckoning Position} \label{subsubsec:dr-position}
Apart from an internally calculated position, the Sphero provides raw sensor data for the accelerometer, gyroscope, orientation and velocity (internally fused from the motor velocity and acceleration). For the scope of this demonstration we make use of this velocity and orientation to compute the position using OpenHPS. The output of this source is shown in Figure~\ref{fig:sensor-individual}c.

\subsubsection{Model Creation} \label{subsubsec:model-creation}
In Listing~\ref{lst:openhps-demo-merge} we combine the four graph shapes for our video output, internal position, input and dead reckoned position. We use a built-in object merging node (lines~22~to~30) that merges frames where the source UID is equal to ``sphero''. The merge node will wait until all of its incoming edges pushed a frame, or the timeout of 20\,ms has been reached. By default, this merge will use the weighted average of all incoming positions, velocities and orientations (with the weight being the inverse of its accuracy). Developers have the choice to choose their own strategy by, for instance, selecting a single position based on the highest accuracy. 

\begin{listing}[htb]
	\begin{openhps}
ModelBuilder.create()
	.addNode(new WorkerNode("video.ts", {
		poolSize: 1,
		name: "video"
	}))
	.addShape(inputSource)
	.addShape(spheroPosition)
	.addShape(spheroVelocity)
	// Feedback loop
	.addShape(GraphBuilder.create()
		.from("merged")
		.debounce(10, TimeUnit.MILLISECOND)
		// Clone the frame and update timestamp
		// (needed to process velocity)
		.clone({
			repack: true
		})
		.via(new VelocityProcessingNode())
		.to("feedback"))
	.addShape(GraphBuilder.create()
		.from("video", "sphero_position", "input", 
			"sphero_velocity", "feedback")
		.merge((frame, options) => options.sourceNode,
			{
				timeout: 20,
				timeoutUnit: TimeUnit.MILLISECOND,
				// Minimum two sources, else the feedback
				// loop will continue
				minCount: 2,
				objectFilter: obj => obj.uid === 'sphero',
			})
		.via("merged")	// Feedback loop
		.to(new CSVDataSink("position.csv", [
			{ id: "timestamp", title: "timestamp" },
			{ id: "x", title: "x" },
			{ id: "y", title: "y" },
		], (frame: DataFrame) => {
			return {
				timestamp: frame.createdTimestamp,
				x: frame.source.getPosition().toVector3().x,
				y: frame.source.getPosition().toVector3().y,
			};
		})))
	.build().then(model => {
		// Model created
	});
	\end{openhps}
	\vspace{-0.2cm}
	\caption{Demonstration model creation}
	\label{lst:openhps-demo-merge}
\end{listing}

This final fused position is presented in Figure~\ref{fig:sensor-merged}. Compared to the individual positioning methods shown in Figure~\ref{fig:sensor-individual}, we have more data points for our positions. This is because we do not wait for all sources to provide data before computing the next position ($20\,ms$ timeout). Our feedback loop called ``feedback'' ensures that position fusion never relies on just one source.

\subsubsection{Evaluation}
We have shown our completed positioning system in the previous section. Four sources and a feedback loop resulted in a fused position. In order to evaluate this positioning model, we removed parts of our video source to simulate an obstacle or blind spots for the camera.

The goal of this evaluation is to first ensure that the positioning model can function with missing information and to determine the error as a result of this missing positioning data.

\begin{figure*}[ht!]
	\centering
	\begin{subfigure}[b]{0.49\textwidth}
		\includegraphics[width=\textwidth, keepaspectratio, page=1]{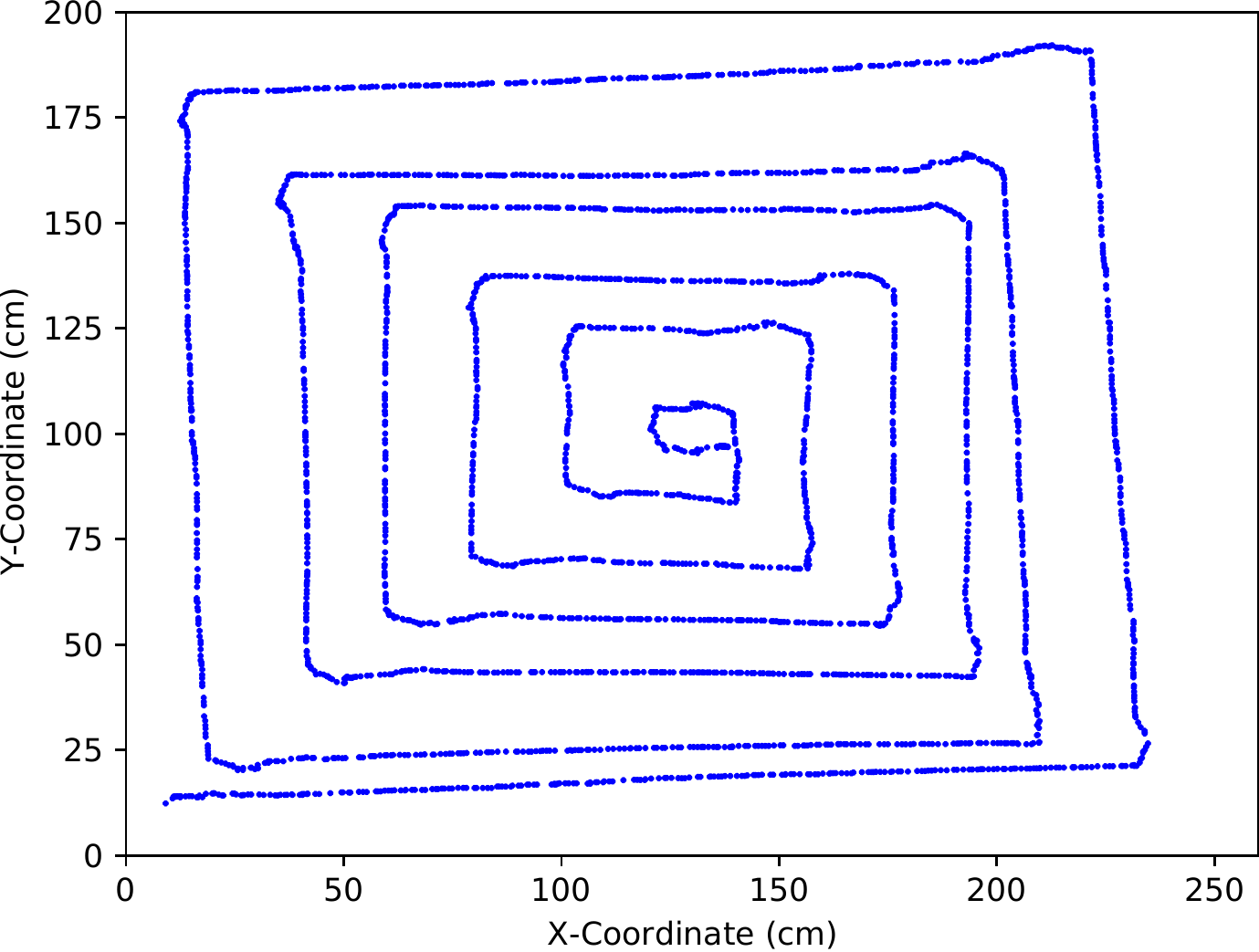}
		\caption{Fused position using all sources} \label{fig:sensor-merged}
		\Description{The plot shows the fused position of all positioning methods. Similar to the video position, the outer most rectangular spirals are at an angle. Corners are more ``precise'' than those shown in the video position, indicating the influence of the input, Sphero position and dead reckoning sources.}
	\end{subfigure}
	\hfill
	\begin{subfigure}[b]{0.49\textwidth}  
		\includegraphics[width=\textwidth, keepaspectratio, page=1]{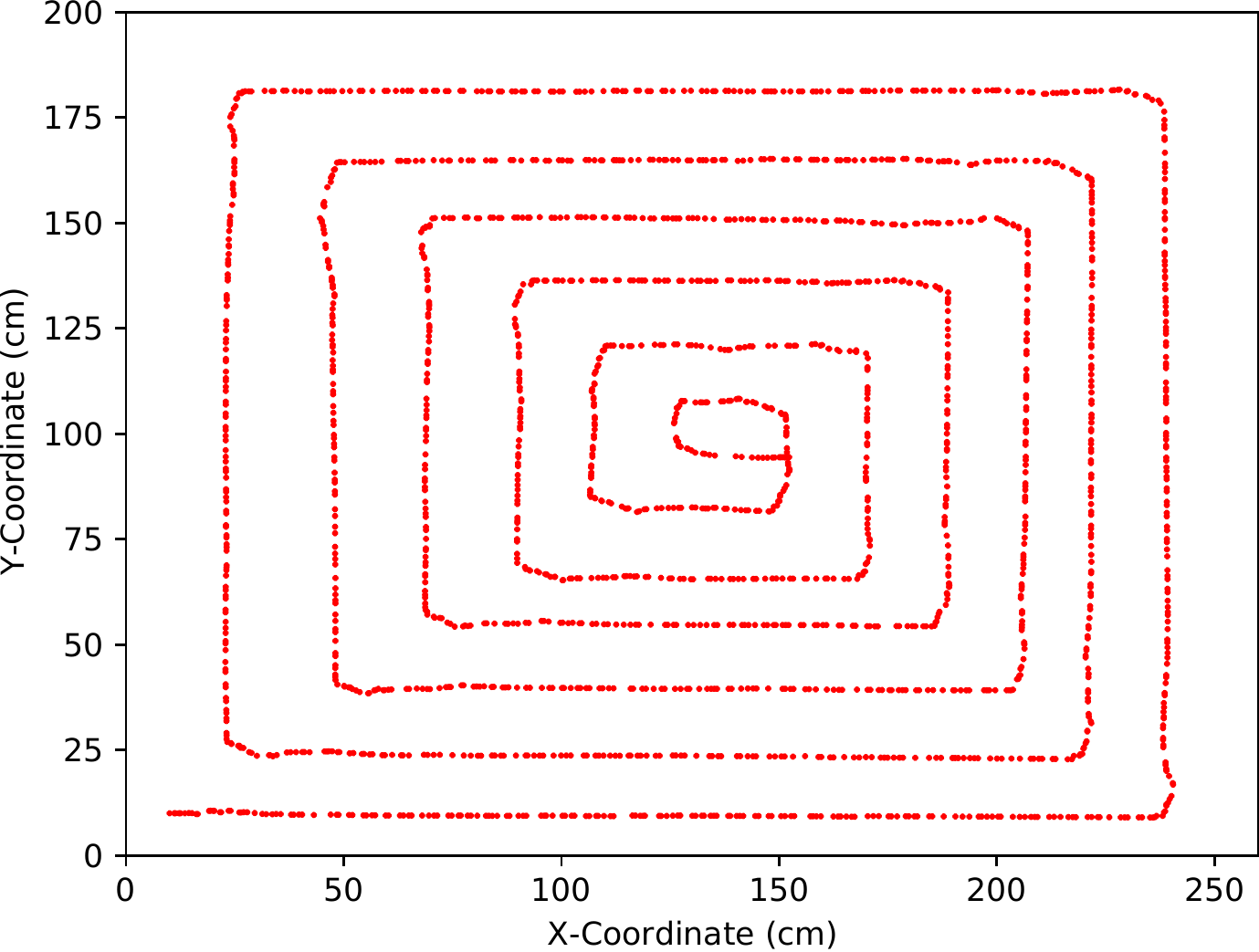}
		\caption{Fused position without video source} \label{fig:sensor-novideo}
		\Description{The second plot shows the fused position without the video source. As expected, the observed angle that is present with the video source is not shown. The X and Y axis are larger than the Sphero position and dead reckoning position, showing the influence of the input source.}
	\end{subfigure}
	\vskip\baselineskip
	\begin{subfigure}[b]{0.49\textwidth}
		\includegraphics[width=\textwidth, keepaspectratio, page=1]{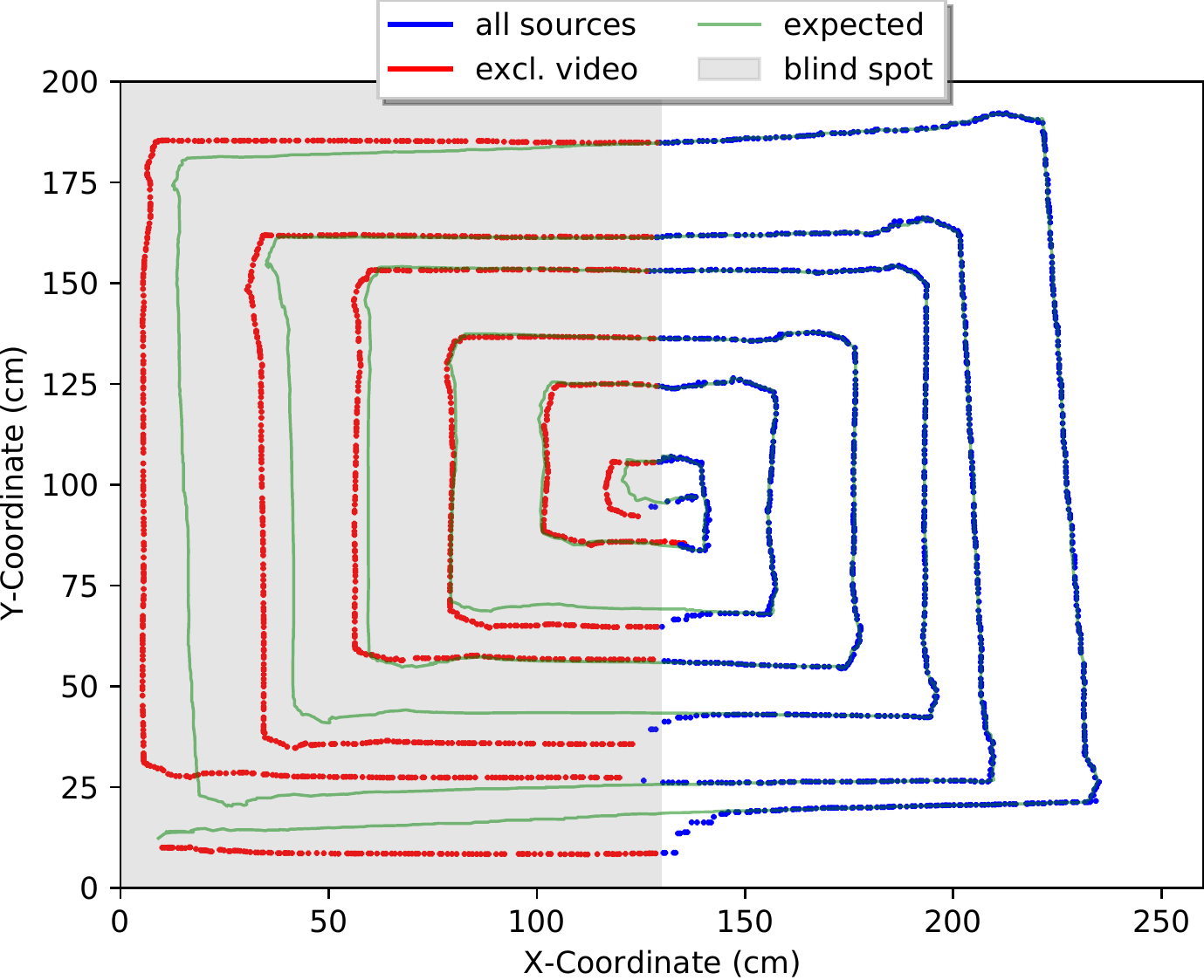}
		\caption{Fused position with camera blind spot on the left} \label{fig:sensor-deadzone-left}
		\Description{In the first evaluation plot, a grey area is shown on the left to indicate a blind spot in the camera. The position starts similar as the fused position without video source. This causes an error from our expected position, as the angle from our video source is not present. Once the Sphero moves past the blind spot, the position gradually moves to that position. This behavior repeats itself for the remaining trajectory.}
	\end{subfigure}
	\hfill
	\begin{subfigure}[b]{0.49\textwidth}  
		\includegraphics[width=\textwidth, keepaspectratio, page=1]{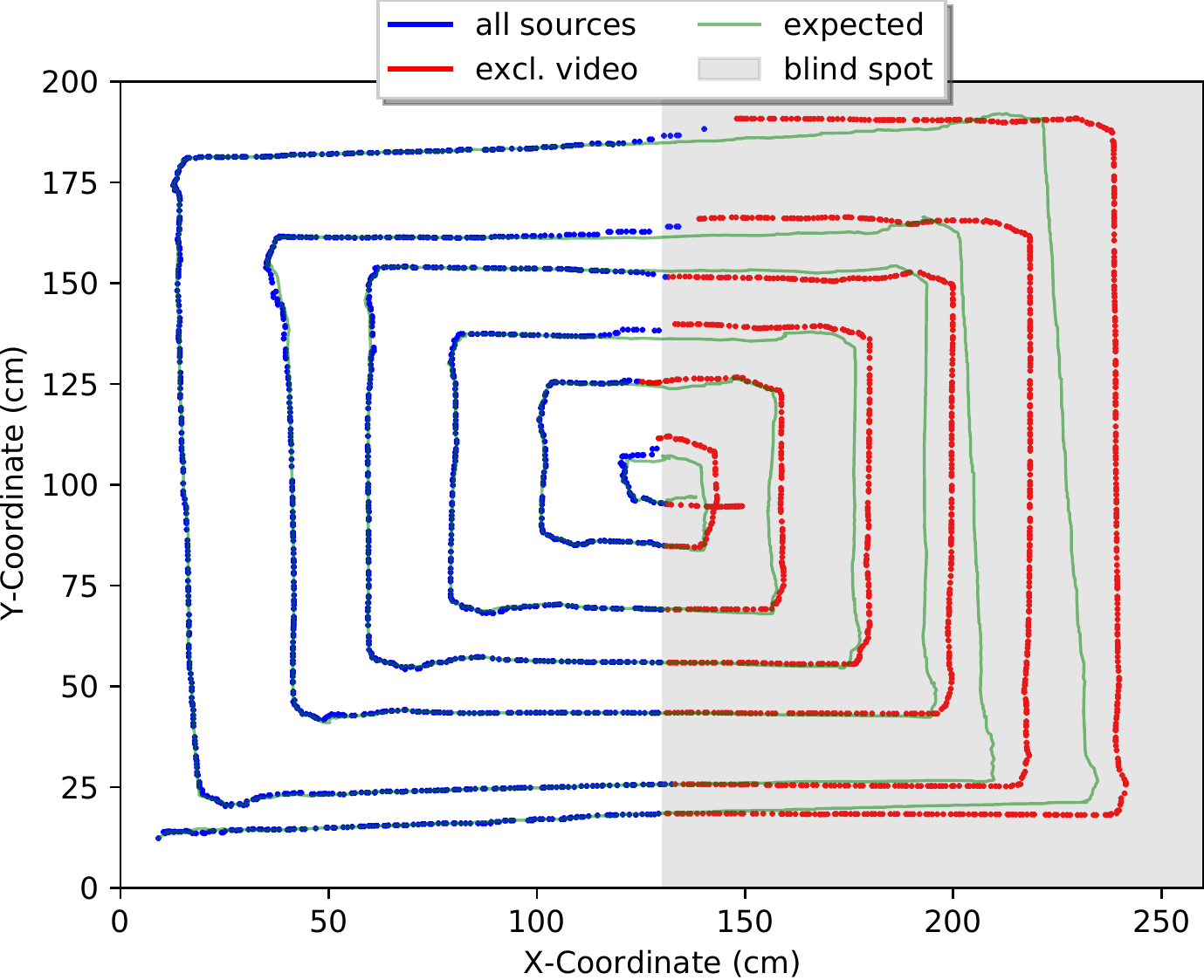}
		\caption{Fused position with camera blind spot on the right} \label{fig:sensor-deadzone-right}
		\Description{In the second evaluation plot, a grey area is shown on the right to indicate a blind spot in the camera. The position starts similar as the fused position (including video source). Once the Sphero reaches the blind spot, the trajectory becomes more straight as the angle of the video source is lost.}
	\end{subfigure}
	\caption{Fused positions processed by our model}\label{fig:sensor-global}
	\vspace{0.4cm}
\end{figure*}

To illustrate a baseline of the remaining sources that will take over the positioning, we show the merged position of all sources except the video source in Figure~\ref{fig:sensor-novideo}.

Figures~\ref{fig:sensor-deadzone-left}~and~\ref{fig:sensor-deadzone-right} show two examples with video blind spots (grey areas). Indicated in blue are the data points where the video processing was still able to detect an object, whereas the positions calculated without input from the video source are highlighted in red in the figures. 

\begin{table}[htb]
	\begin{tabular}{c|c|c}
		\textbf{Source(s)} & \textbf{Avg error} & \textbf{Max error} \\ \hline
		all sources (Fig.~\ref{fig:sensor-merged}) & 0.00\,cm & 0.00\,cm \\
		input control only (Fig.~\ref{fig:sensor-individual}a) & 23.07\,cm & 50.06\,cm \\
		internal position only (Fig.~\ref{fig:sensor-individual}b) & 16.16\,cm & 33.38\,cm \\
		dead reckoning only (Fig.~\ref{fig:sensor-individual}c) & 17.09\,cm & 34.44\,cm \\
		video source only (Fig.~\ref{fig:sensor-individual}d) & 1.30\,cm & 4.74\,cm \\
		all sources excl. video (Fig.~\ref{fig:sensor-novideo}) & 13.59\,cm & 29.73\,cm \\
		blind spot left (Fig.~\ref{fig:sensor-deadzone-left}) & 4.26\,cm & 21.65\,cm \\
		blind spot right (Fig.~\ref{fig:sensor-deadzone-right}) & 4.81\,cm & 24.40\,cm \\
	\end{tabular}
	\vspace{0.1cm}
	\captionof{table}{Average and maximum XY position error compared to the fused position with all sources} \label{table:demo-error}
	\vspace{-0.4cm}
\end{table}

In Table~\ref{table:demo-error}, we show the average and maximum position error compared to the final fused position from Figure~\ref{fig:sensor-merged}. This error is determined by taking 100 timestamped key points in each trajectory (every 51\,ms) and calculating the average and maximum difference for those points.

Our results show that the video source is the main positioning method in the fused position. Blind spots in this source result in the model falling back to the remaining dead reckoning. However, the positioning model is self correcting and will gradually align with the video source position once it becomes available.

The positioning model we illustrated in Figure~\ref{fig:demo-model} is highly adaptable depending on the desired outcome. For example, noise filtering nodes such as a Simple Moving Average~(SMA) can be used on the video accuracy to provide a smoother transition at the border of the blind spot.

With the evaluation in Figure~\ref{fig:sensor-global} and Table~\ref{table:demo-error} we have proven that multiple producers of sensor information can be merged together into a continuous stream of fused positions. By creating blind spots in our video source, we have shown that the model is capable of running without our main visual positioning method.

\section{Conclusion and Future Work}
\label{sec:conclusion}

We have presented OpenHPS, an open source hybrid positioning system. We focused on the different actors of our system that have been defined based on an investigation of some of the more prominent existing positioning methods and algorithms. These actors, in combination with our requirements, were used in developing our positioning framework with its graph topology. We further presented our definition of \emph{nodes}, \emph{data frames}, \emph{data objects} as well as \emph{positions}. Finally, the OpenHPS implementation in TypeScript highlighting how we addressed and satisfied our non-functional requirements, has been discussed in Section~\ref{subsec:implementation}.

In the demonstrator application in Section~\ref{subsec:demonstrator}, we have illustrated how multiple positioning methods can be fused via some high-level decision fusion. We have further highlighted---by removing certain parts of our main sensor source---how the presented positioning model continues to work on the remaining positioning methods and manages to recover once the input from the main sensor source is back.

A major effort in the design and development of OpenHPS went into the extensibility of our framework. External modules can be used to extend OpenHPS with additional positioning methods and techniques. Some basic positioning methods are currently included in the core OpenHPS component. However, in order to prevent that the core contains potentially unused nodes, in the future some of these basic positioning methods and algorithms might be moved to their own dedicated modules (e.g.~for fingerprinting techniques). Apart from individual nodes, these modules can also provide complete graph shapes that act similar to position providers in other high-level hybrid positioning systems.

The real-time processing of positioning information was the most important goal for the presented OpenHPS framework. During the development, the computing performance of the positioning model has therefore always received a high priority and lead to the introduction of worker nodes and services. Future research and development of the OpenHPS hybrid positioning system might focus on optimising the serialisation of data frames in order to only serialise changes in data rather than all available information. This optimisation would ensure that data transfers are limited to new data only.

Overall, the presented OpenHPS framework represents a solid hybrid positioning solution offering various possibilities for future extensions. An obvious future extension would be the introduction of additional layers of abstraction providing similar high-level functionality as offered by some of the solutions discussed in the related work. Further, the exiting reference spaces might be extended in a separate OpenHPS module in order to represent and deal with symbolic locations, similar as offered by HyLocSys~\cite{ficco2009hybrid}. The support of symbolic locations~\cite{hightower2001location} will further strengthen the position of OpenHPS as a framework for context-aware computing and implicit human-computer interaction.

\bibliographystyle{ACM-Reference-Format}
\balance
\bibliography{openHPS}

\end{document}